\documentclass[11pt, a4paper]{lumia}
\pdfoutput=1

\usepackage[sort&compress]{natbib}
\usepackage{xspace}

\theoremstyle{plain}

\newtheorem*{proposition*}{Proposition}

\theoremstyle{definition}

\theoremstyle{definition}

\def\eqref#1{equation~\ref{#1}}

\usepackage{graphicx}           
\usepackage{tikz}               
\usepackage[edges]{forest}      

\usepackage{url}                
\usepackage{xurl}               

\usepackage{array}              
\usepackage{longtable}          
\usepackage{multirow}           
\usepackage{makecell}           
\usepackage{ragged2e}           

\usepackage{mathtools}          
\usepackage{nicefrac}           

\usepackage{algorithm}          
\usepackage{algorithmicx}       
\usepackage{algpseudocode}      
\usepackage{listings}           

\usepackage{subcaption}         
\usepackage{wrapfig}            
\usepackage[export]{adjustbox}  

\usepackage{changes}            
\usepackage{xspace}             
\usepackage[normalem]{ulem}     
\usepackage{CJKutf8}            

\usepackage[tikz]{bclogo}       
\usepackage[framemethod=tikz]{mdframed} 

\usepackage{placeins}
\usepackage{cleveref}
\usepackage{lipsum}             
\usepackage{tocloft}            
\usepackage{afterpage}          
\usepackage{bbding}             
\usepackage{epigraph}           
\usepackage{minitoc}            
\usepackage{multicol}           
\usepackage{textgreek}          

\newcommand{\hlfirst}[1]{\begingroup\setlength{\fboxsep}{1pt}\colorbox[HTML]{CFE2FF}{#1}\endgroup}
\newcommand{\hlsecond}[1]{\begingroup\setlength{\fboxsep}{1pt}\colorbox[HTML]{E8F1FF}{#1}\endgroup}

\definecolor{mygreen}{rgb}{0.29, 0.7, 0.48}
\definecolor{darksalmon}{rgb}{0.91, 0.59, 0.48}
\definecolor{mygrey}{gray}{0.4}

\usepackage[most]{tcolorbox}
\usepackage{makecell}
\newcolumntype{P}[1]{>{\RaggedRight\arraybackslash}p{#1}}

\definecolor{uclablue}{RGB}{39, 116, 174}
\definecolor{bigaired}{RGB}{156, 0, 0}
\definecolor{myblue}{HTML}{598BE7}
\definecolor{mildblue}{RGB}{31,119,180}
\definecolor{sectionblue}{RGB}{70, 130, 180}
\definecolor{methodblue}{RGB}{0, 150, 136}
\definecolor{bgblue}{RGB}{245,243,253}
\definecolor{ttblue}{RGB}{91,194,224}
\definecolor{mygreen}{rgb}{0.64, 0.56, 0.88}
\definecolor{myyellow}{rgb}{0.68, 0.6, 0.1}
\definecolor{fancygreen}{rgb}{0.33, 0.68, 0.20}
\definecolor{salmon}{rgb}{0.94, 0.52, 0.49}
\definecolor{tablegreen}{rgb}{0.82, 0.94, 0.75}
\definecolor{tableblue}{rgb}{0.81, 0.90, 0.94}
\definecolor{tablered}{rgb}{0.97, 0.85, 0.85}
\definecolor{tableorange}{rgb}{0.96, 0.85, 0.81}
\definecolor{myorange}{rgb}{1.0, 0.49, 0.0}
\definecolor{tlgreen}{rgb}{0.33, 0.68, 0.20}
\definecolor{darkgreen}{RGB}{0,100,0}
\definecolor{darkred}{RGB}{200, 0, 0}
\definecolor{customyellow}{HTML}{FFFACD}
\definecolor{refinegreen}{RGB}{0, 128, 75}
\definecolor{scoregreen}{RGB}{34, 139, 34}
\definecolor{hidden-blue}{RGB}{194,232,247}
\definecolor{hidden-black}{RGB}{20,68,106}
\definecolor{yes}{HTML}{C6EFCE}
\definecolor{no}{HTML}{FFC7CE}
\definecolor{partial}{HTML}{FFEB9C}
\definecolor{external}{HTML}{D9E1F2}
\definecolor{hdr}{HTML}{F2F2F2}
\definecolor{GRPOrow}{gray}{0.96}
\definecolor{FlowRLrow}{RGB}{225,236,255}
\definecolor{FlowBlue}{RGB}{80,120,210}
\definecolor{GRPOGray}{gray}{0.35}

\setlist[itemize]{leftmargin=20pt, noitemsep, topsep=0pt}

\NewDocumentCommand{\kaiyan}{mO{}}{\textcolor{purple}{\textsuperscript{\textit{kaiyan}}\textsf{\textbf{\small[#1]}}}}
\NewDocumentCommand{\yuxin}{mO{}}{\textcolor{cyan}{\textsuperscript{\textit{yuxin}}\textsf{\textbf{\small[#1]}}}}
\NewDocumentCommand{\bx}{mO{}}{\textcolor{green}{\textsuperscript{\textit{bx}}\textsf{\textbf{\small[#1]}}}}
\NewDocumentCommand{\at}{mO{}}{\textcolor{red}{\textsuperscript{\textit{AT}}\textsf{\textbf{\small[#1]}}}}
\NewDocumentCommand{\re}{mO{}}{\textcolor{blue}{\textsuperscript{\textit{RE}}\textsf{\textbf{\small[#1]}}}}
\NewDocumentCommand{\ybsun}{mO{}}{\textcolor{magenta}{\textsuperscript{\textit{youbang}}\textsf{\textbf{\small[#1]}}}}
\NewDocumentCommand{\runze}{mO{}}{\textcolor{orange}{\textsuperscript{\textit{runze}}\textsf{\textbf{\small[#1]}}}}
\NewDocumentCommand{\add}{mO{}}{\textcolor{darkgreen}{\textsuperscript{\textit{Maybe Consider Discuss}}\textsf{\textbf{[#1]}}}}

\newcommand{\cmark}{\textcolor{darkgreen}{\boldmath$\checkmark$}}
\newcommand{\xmark}{\textcolor{darkred}{\boldmath$\times$}}

\newenvironment{itemize*}%
 {\leftmargini=10pt\begin{itemize}%
  \setlength{\itemsep}{0pt}%
  \setlength{\parskip}{0pt}%
  }%
 {\end{itemize}}

\newenvironment{enumerate*}%
 {\begin{enumerate}%
  \setlength{\itemsep}{0pt}%
  \setlength{\parskip}{0pt}}%
 {\end{enumerate}}

\newcommand{\cellstatus}[1]{%
  \begingroup
  \StrTrim{#1}[\statusval]%
  \IfStrEq{\statusval}{Yes}{\cellcolor{yes}\cmark}{}%
  \IfStrEq{\statusval}{No}{\cellcolor{no}\xmark}{}%
  \IfBeginWith{\statusval}{Yes (}{\cellcolor{yes}\cmark~\textit{\statusval\unskip}}{}%
  \IfStrEq{\statusval}{Partial}{\cellcolor{partial}\textbf{Partial}}{}%
  \IfStrEq{\statusval}{External}{\cellcolor{external}\textbf{External}}{}%
  \endgroup
}

\newtcolorbox{myboxi}[1][]{
  breakable,
  title=#1,
  colback=red!5,
  colbacktitle=red!5,
  coltitle=black,
  fonttitle=\bfseries,
  bottomrule=0pt,
  toprule=0pt,
  leftrule=2pt,
  rightrule=2pt,
  titlerule=0pt,
  arc=0pt,
  outer arc=0pt,
  colframe=red,
}

\newtcolorbox{myboxnote}[1][]{
  breakable,
  title=#1,
  colback=orange!0,
  colbacktitle=orange!0,
  coltitle=black,
  fonttitle=\bfseries,
  bottomrule=0pt,
  toprule=0pt,
  leftrule=2pt,
  rightrule=2pt,
  titlerule=0pt,
  arc=0pt,
  outer arc=0pt,
  colframe=orange,
}

\newtcolorbox{myboxii}[1][]{
  breakable,
  freelance,
  title=#1,
  colback=white,
  colbacktitle=white,
  coltitle=black,
  fonttitle=\bfseries,
  bottomrule=0pt,
  boxrule=0pt,
  colframe=white,
  overlay unbroken and first={
  \draw[red!75!black,line width=3pt]
    ([xshift=5pt]frame.north west) -- 
    (frame.north west) -- 
    (frame.south west);
  \draw[red!75!black,line width=3pt]
    ([xshift=-5pt]frame.north east) -- 
    (frame.north east) -- 
    (frame.south east);
  },
  overlay unbroken app={
  \draw[red!75!black,line width=3pt,line cap=rect]
    (frame.south west) -- 
    ([xshift=5pt]frame.south west);
  \draw[red!75!black,line width=3pt,line cap=rect]
    (frame.south east) -- 
    ([xshift=-5pt]frame.south east);
  },
  overlay middle and last={
  \draw[red!75!black,line width=3pt]
    (frame.north west) -- 
    (frame.south west);
  \draw[red!75!black,line width=3pt]
    (frame.north east) -- 
    (frame.south east);
  },
  overlay last app={
  \draw[red!75!black,line width=3pt,line cap=rect]
    (frame.south west) --
    ([xshift=5pt]frame.south west);
  \draw[red!75!black,line width=3pt,line cap=rect]
    (frame.south east) --
    ([xshift=-5pt]frame.south east);
  },
}

\tcbset{
  takeawaysbox/.style={
    title=Takeaways,
    colback=lightblue!80,
    colframe=black,
    fonttitle=\bfseries\small,
    coltitle=white,
    colbacktitle=black,
    enhanced,
    attach boxed title to top left={xshift=2.5mm,yshift=-2.5mm},
    boxed title style={rounded corners, size=small, colframe=black, colback=black},
    width=\linewidth,
    arc=3.5mm
  }
}

\mdfdefinestyle{mystyle}{%
  rightline=true,
  innerleftmargin=10,
  innerrightmargin=10,
  outerlinewidth=3pt,
  topline=false,
  rightline=true,
  bottomline=false,
  skipabove=\topsep,
  skipbelow=\topsep
}

\tikzset{%
    every node/.style={font=\tiny},
    parent/.style =          {align=center,text width=2cm,rounded corners=3pt, line width=0.3mm, fill=gray!10,draw=gray!80},
    child/.style =           {align=center,text width=2.0cm,rounded corners=3pt, fill=blue!10,draw=blue!80,line width=0.3mm},
    grandchild/.style =      {align=center,text width=2cm,rounded corners=3pt},
    greatgrandchild/.style = {align=center,text width=1.5cm,rounded corners=3pt},
    greatgrandchild2/.style = {align=center,text width=1.5cm,rounded corners=3pt},    
    referenceblock/.style =  {align=center,text width=1.5cm,rounded corners=2pt},
    pretrain/.style =           {align=center,text width=2.0cm,rounded corners=3pt, fill=blue!10,draw=blue!80,line width=0.3mm},   
    pretrain_work/.style =           {align=center, text width=8.5cm,rounded corners=3pt, fill=blue!10,draw=blue!0,line width=0.3mm},  
    template/.style =           {align=center,text width=2.0cm,rounded corners=3pt, fill=red!10,draw=red!80,line width=0.3mm},   
    template_work/.style =           {align=center,text width=8.5cm,rounded corners=3pt, fill=red!10,draw=red!0,line width=0.3mm},    
    answer/.style =           {align=center,text width=2.0cm,rounded corners=3pt, fill= cyan!10,draw= cyan!80,line width=0.3mm},   
    answer_work/.style =           {align=center,text width=8.5cm,rounded corners=3pt, fill= cyan!10,draw= cyan!0,line width=0.3mm},      
    multiple/.style =           {align=center,text width=2.0cm,rounded corners=3pt, fill= orange!10,draw= orange!80,line width=0.3mm},   
    multiple_work/.style =           {align=center,text width=8.5cm,rounded corners=3pt, fill= orange!10,draw= orange!0,line width=0.3mm},        
    tuning/.style =           {align=center,text width=2.0cm,rounded corners=3pt, fill= magenta!10,draw= magenta!80,line width=0.3mm},   
    tuning_work/.style =           {align=center,text width=8.5cm,rounded corners=3pt, fill= magenta!10,draw= magenta!0,line width=0.3mm},          
}

\newcommand{\lstbg}[3][0pt]{{\fboxsep#1\colorbox{#2}{\strut #3}}}

\lstdefinelanguage{diff}{
  basicstyle=\ttfamily\small,
  morecomment=[f][\lstbg{red!20}]-,
  morecomment=[f][\lstbg{green!20}]+,
}

\lstdefinelanguage{diffpython}{
  language=diff,
  morekeywords={def, if, else, for, while, return, import, from, as, class, with, try, except, finally, raise, lambda, and, or, not, in, is, None, True, False},
  morecomment=[l]{\#},
  morestring=[b]",
  morestring=[b]',
}

\usepackage{dsfont}
\usepackage{fvextra}
\usepackage{fancyvrb}

\setheadertext{
  \includegraphics[height=1.2cm]{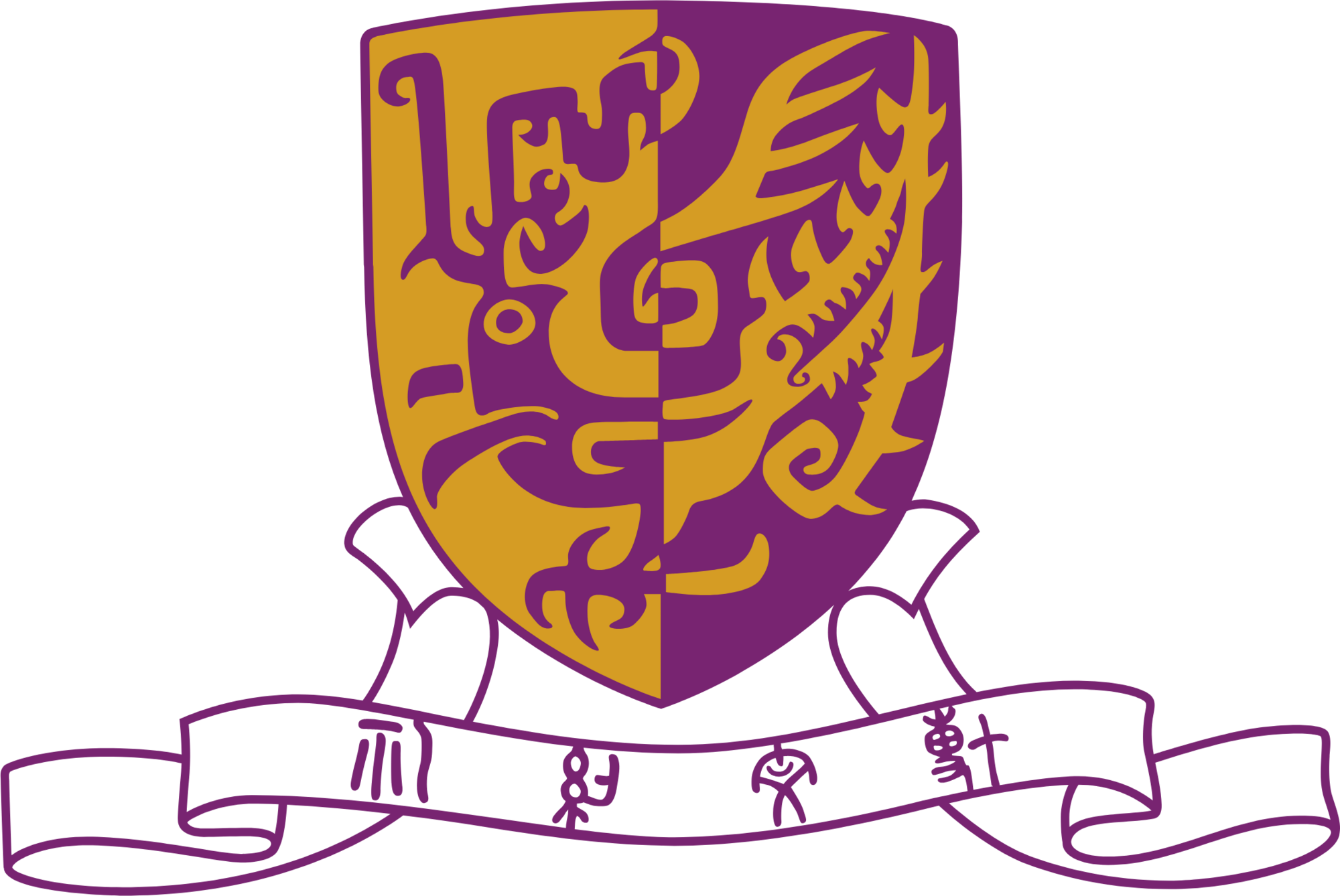} \hspace{0.1cm}
  \includegraphics[height=1.2cm]{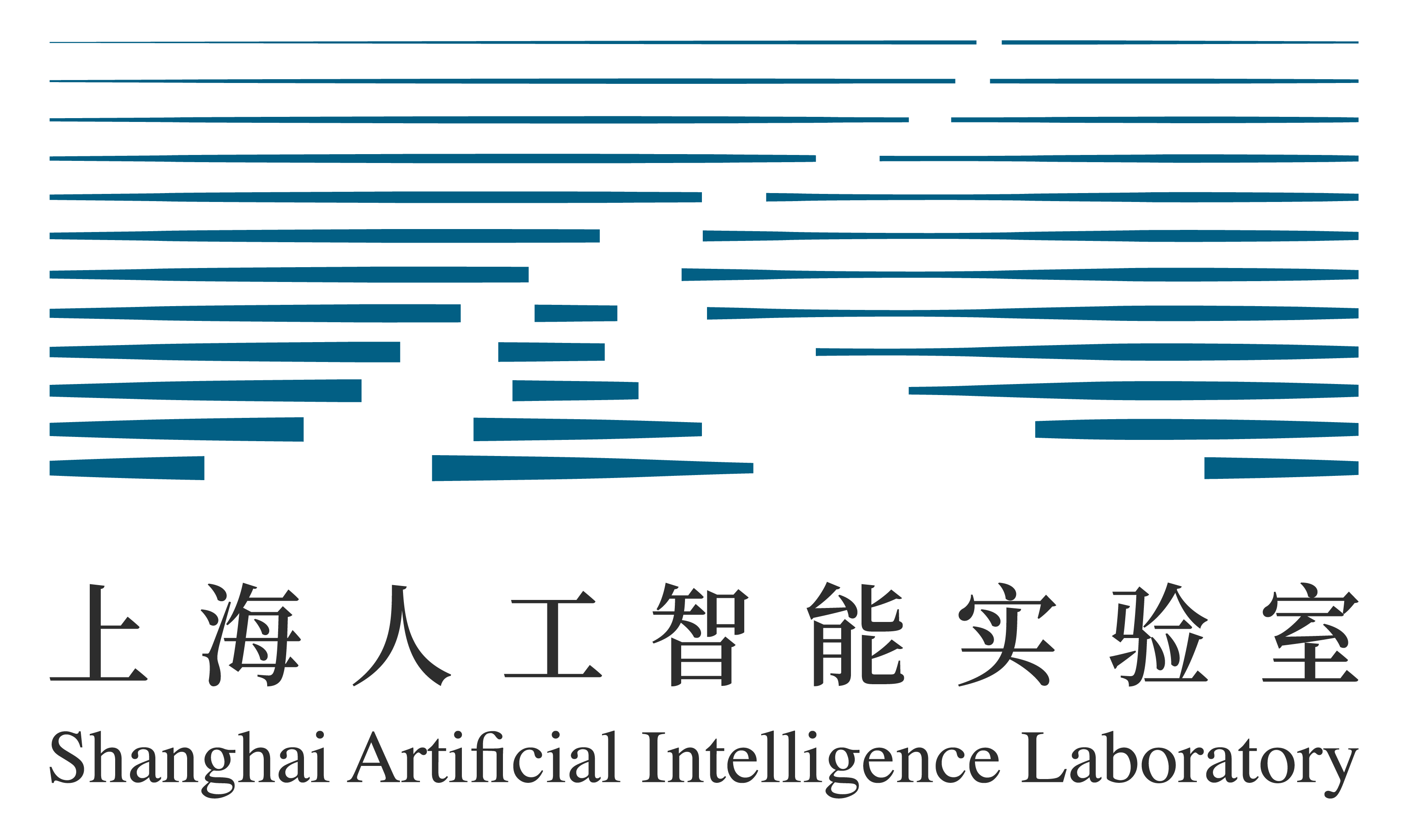} \hspace{0.1cm}
  \includegraphics[height=1.2cm]{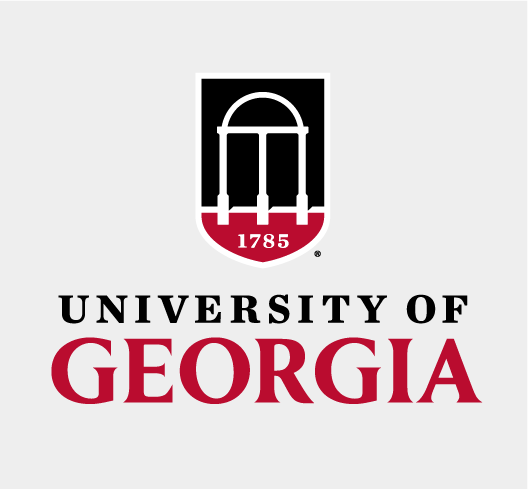} \hspace{0.1cm}
  \includegraphics[height=1.2cm]{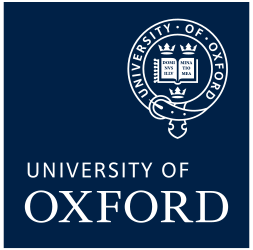}
}

\correspondingemail{
  {\emailicon\ }\href{mailto:xxythecoder@gmail.com}{xxythecoder@gmail.com} \quad $^\ast$Equal contribution. \quad $^\dagger$Corresponding authors.
}

\githublink{https://github.com/xxyQwQ/StraTA}

\huggingfacelink{https://huggingface.co/collections/xxyQwQ/strata}

\renewcommand{\today}{2026-05-06}

\title{StraTA: Incentivizing Agentic Reinforcement Learning with Strategic Trajectory Abstraction}
\setheadertitle{StraTA: Incentivizing Agentic Reinforcement Learning with Strategic Trajectory Abstraction}

\author{
  Xiangyuan Xue$^{1,2\ast}$, Yifan Zhou$^{3\ast}$, Zidong Wang$^{1}$, Shengji Tang$^{1}$, \newline
  Philip Torr$^{4}$, Wanli Ouyang$^{1,2,5\dagger}$, Lei Bai$^{2\dagger}$, Zhenfei Yin$^{4\dagger}$ \\
  $^1$The Chinese University of Hong Kong \quad $^2$Shanghai Artificial Intelligence Laboratory \newline
  $^3$University of Georgia \quad $^4$University of Oxford \quad $^5$Shenzhen Loop Area Institute
}

\begin{document}

\begin{abstract}
Large language models (LLMs) are increasingly used as interactive agents, but optimizing them for long-horizon decision making remains difficult because current methods are largely purely reactive, which weakens both exploration and credit assignment over extended trajectories. In this work, we present \textbf{Stra}tegic \textbf{T}rajectory \textbf{A}bstraction (StraTA), a simple framework that introduces an explicit trajectory-level strategy into agentic reinforcement learning (RL). StraTA samples a compact strategy from the initial task state, conditions subsequent actions on that strategy, and trains strategy generation and action execution jointly with a hierarchical GRPO-style rollout design, further enhanced by diverse strategy rollout and critical self-judgment. Experiments on ALFWorld, WebShop, and SciWorld show that StraTA consistently improves both sample efficiency and final performance over strong baselines. StraTA reaches success rates of 93.1\% on ALFWorld and 84.2\% on WebShop. On SciWorld, StraTA attains a 63.5\% overall score, outperforming frontier closed-source models.
\end{abstract}

\maketitle

\section{Introduction}
\label{sec:introduction}

Large language models (LLMs) are increasingly deployed as interactive agents that perceive environments, reason in natural language, and act through external tools or APIs. This paradigm has enabled rapid progress on complex tasks such as web interaction, computer operation, and scientific problem solving~\citep{xi2025rise,wang2024survey}. Nevertheless, strong general-purpose capabilities do not automatically translate into reliable performance on specialized long-horizon tasks. Agents must often recover from earlier mistakes, maintain coherent behavior over many turns, and make decisions whose consequences may only become apparent much later in the trajectory.

\begin{figure}[t]
    \centering
    \includegraphics[width=1.0\textwidth]{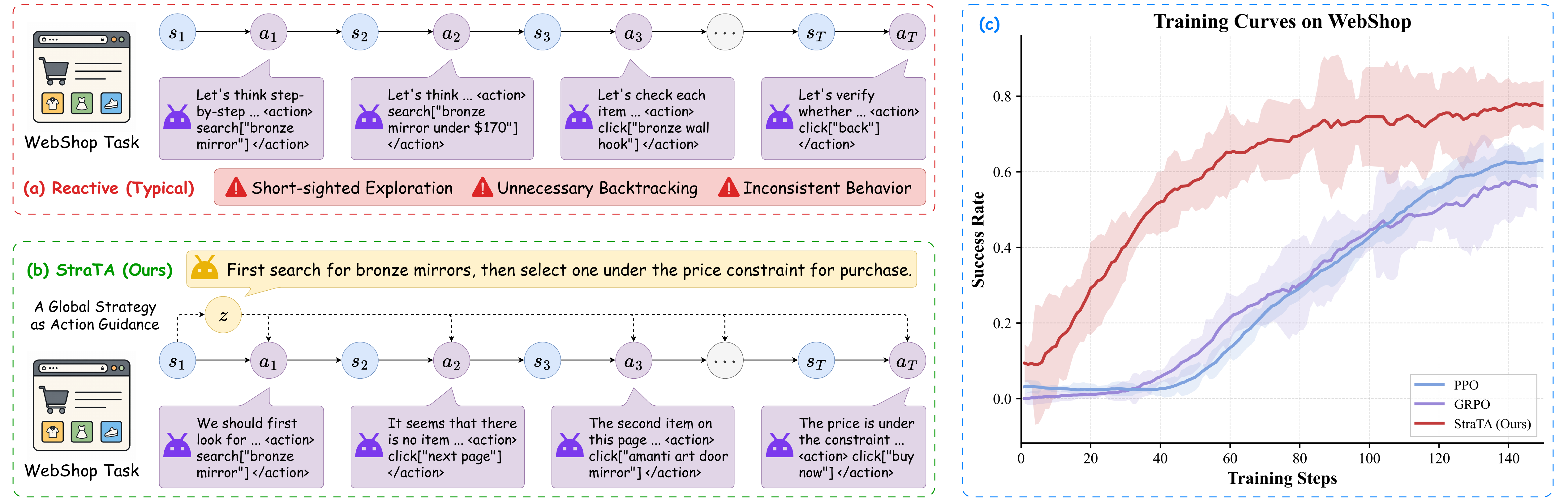}
    \caption{A comparison between reactive agent and our proposed StraTA framework. (a) Reactive agent generates actions from local states alone, planning and executing at the same time, which may lead to short-sighted exploration, unnecessary backtracking, and inconsistent behavior. (b) StraTA framework first generates a compact global strategy and conditions all subsequent actions on this, enabling more coherent and consistent execution. (c) On long-horizon agentic tasks, explicit strategy guidance significantly improves learning efficiency and final performance.}
    \label{fig:teaser}
\end{figure}

Agentic reinforcement learning (RL) provides a natural way to improve such behavior by optimizing agents from environment feedback~\citep{zhang2025landscape}. Compared with conventional language modeling, however, this setting is particularly challenging. Each decision is hierarchical: the policy must choose a step-level action while autoregressively generating the token sequence that realizes it. At the same time, rewards are often sparse and delayed until the episode terminates. Together, these properties make long-horizon exploration difficult and substantially complicate credit assignment.

Recent methods have made encouraging progress. PPO-style and GRPO-style training have shown that RL can improve multi-turn LLM agents~\citep{xi2025agentgymrl,rllm2025}, while newer variants further refine the learning signal for long-horizon interaction~\citep{zhou2024archer,feng2025group}. However, most existing formulations remain purely reactive. As shown in Figure~\ref{fig:teaser}a, at each step the agent predicts the next action solely from the current state. The policy therefore has to decide both the immediate next move and the overall course of action at the same time. This coupling often yields short-sighted exploration, unnecessary backtracking, and internally inconsistent behavior.

We argue that a key missing abstraction is an explicit trajectory-level strategy~\citep{erdogan2025plan}. Human problem solving is rarely purely reactive: people typically form a high-level plan, execute local actions under that plan, and later reflect on whether the plan was appropriate. Motivated by this observation, we propose \textbf{Stra}tegic \textbf{T}rajectory \textbf{A}bstraction (StraTA), a framework that introduces explicit strategy guidance into agentic RL. Figure~\ref{fig:teaser}b illustrates how StraTA differs from reactive agents. At the beginning of each episode, StraTA samples a strategy from the initial task state, which is then fixed and prepended to every subsequent decision, so each action is generated from both the global strategy and the local state. In this way, long-horizon agentic decision making is decomposed into two objectives: generating a good strategy and executing actions that faithfully follow it.

To optimize these two levels jointly, StraTA builds a hierarchical rollout structure on top of GRPO-style training. For each task, the policy first samples multiple strategies and then performs multiple rollouts under each strategy. This design enables strategy-level comparisons across different plans and action-level comparisons across trajectories generated under the same plan. Beyond trajectory-level outcome rewards, StraTA further assigns each strategy a reward based on the top-performing fraction of its rollouts, yielding a reliable estimate of strategy quality. StraTA also incorporates two simple techniques that further facilitate learning: diverse strategy rollout via farthest-point sampling, which broadens semantic exploration over candidate strategies, and a step-level auxiliary reward via critical self-judgment, which penalizes actions that neither follow the strategy nor advance task progress.

Experiments on multiple agentic benchmarks show that StraTA improves both sample efficiency and final task performance. On ALFWorld~\citep{shridhar2021alfworld} and WebShop~\citep{yao2022webshop}, StraTA consistently outperforms both prompting-based agents and RL methods across model scales: with a 1.5B backbone, it reaches 90.7\% success on ALFWorld and 82.5\% on WebShop, surpassing GiGPO~\citep{feng2025group} by up to 4.0\% and 17.5\%, respectively, and with a 7B backbone, it further improves to 93.1\% and 84.2\%. On the more challenging SciWorld~\citep{wang2022scienceworld} benchmark, StraTA attains a 63.5\% overall score, outperforming both frontier closed-source models and prior RL baselines by 6.1\% and 6.5\%, respectively, while achieving a perfect 100.0\% score on the Lifespan subset. In summary, our contributions are threefold: (1) we introduce StraTA, a strategy-guided framework that equips agentic RL with explicit trajectory-level abstraction; (2) we develop a hierarchical GRPO-style training with top-performing strategy rewards, diverse strategy rollout, and critical self-judgment to improve exploration and credit assignment; and (3) we validate its effectiveness through extensive experiments on long-horizon agentic tasks.

\section{Related Work}
\label{sec:related_work}

\noindent \textbf{LLM-based agents.} LLM-based agents augment foundation models with external tools, such as search engines and code interpreters, enabling them to address complex real-world tasks~\citep{xi2025rise,wang2024survey}. Early work primarily relied on prompting-based designs to build runnable agents for relatively simple and specialized tasks~\citep{yao2023react,wang2024executable,schick2023toolformer,shinn2023reflexion}. More recent studies introduce training-based methods to improve performance on more challenging and general tasks~\citep{zeng2024agenttuning,song2024agentbank,xi2025agentgym,xi2025agentgymrl}. A parallel line of work investigates multi-agent collaboration as a way to further enhance task performance~\citep{hong2024metagpt,wu2024autogen,du2024improving,zhuge2024gptswarm,xue2025comas}. These agents have now been applied to a wide range of domains, from repetitive short-horizon routines to more autonomous long-horizon settings, including computer operation~\citep{yao2022webshop,zhou2024webarena,xie2024osworld}, embodied manipulation~\citep{shridhar2021alfworld,wang2023voyager,chen2025robotwin}, and scientific discovery~\citep{wang2022scienceworld,lu2024ai,xu2025probing}. In this work, we study how RL can further extend the capabilities of LLM-based agents on difficult long-horizon tasks.

\noindent \textbf{Reinforcement fine-tuning.} RL was first introduced to LLMs mainly for alignment with human preferences, a paradigm widely known as reinforcement learning from human feedback (RLHF)~\citep{ziegler2019fine,ouyang2022training}. Most early methods in this line are based upon proximal policy optimization (PPO)~\citep{schulman2017proximal}, with later variants improving training stability and efficiency~\citep{rafailov2023direct,azar2024general,ethayarajh2024kto,ahmadian2024back,hu2025reinforce}. More recently, reinforcement learning with verifiable rewards (RLVR) has become a popular paradigm for improving reasoning, where group relative policy optimization (GRPO)~\citep{shao2024deepseekmath} serves as a representative foundation and has demonstrated strong scalability on complex reasoning tasks~\citep{guo2025deepseek}. Many follow-up methods further improve its effectiveness~\citep{liu2025understanding,lin2025cppo,yu2025dapo,zheng2025group}. Another direction explores RL with intrinsic reward signals in the absence of ground-truth labels~\citep{yuan2024self,prabhudesai2025maximizing,zhao2025learning,zhang2025right,zuo2025ttrl}, which also shows promising results. In this work, we adopt a GRPO-style training framework, but focus on the more challenging setting of long-horizon agentic tasks rather than single-turn reasoning problems.

\noindent \textbf{Agentic reinforcement learning.} Agentic RL is a distinct emerging area in which agents learn long-horizon interactive behavior directly through trial and error in an environment~\citep{zhang2025landscape}. Compared with conventional LLM-based agents, it requires online exploration and adaptation rather than the execution of a fixed routine. Compared with reinforcement fine-tuning, it places greater emphasis on long-horizon decision-making instead of single-turn reasoning. Representative tasks span computer operation~\citep{yao2022webshop,zhou2024webarena,xie2024osworld}, interactive gameplay~\citep{wang2023voyager,zhai2024fine,xi2025agentgym,xi2025agentgymrl}, and embodied manipulation~\citep{shridhar2021alfworld,wang2022scienceworld}. A growing body of work aims at addressing these challenges. Beyond efforts that establish training frameworks~\citep{luo2025agent,feng2025group,rllm2025}, one research direction focuses on improving active exploration in multi-turn interaction~\citep{putta2024agent,wang2025ragen,feng2025towards}, while another studies more accurate and fine-grained credit assignment~\citep{zhou2024archer,liu2025agentic,ji2025tree,feng2025group,he2026hierarchy,peng2026hiper}. Our proposed StraTA framework supports both goals through a simple but effective design that introduces an explicit trajectory-level strategy to guide the entire episode.

\section{Preliminaries}
\label{sec:preliminaries}

\noindent \textbf{Long-horizon agentic tasks.} We formulate a long-horizon agentic task as a finite-horizon Markov decision process (MDP)~\citep{zhang2025landscape}, $(\mathcal{S}, \mathcal{A}, P, R, H, \gamma)$, where $\mathcal{S}$ and $\mathcal{A}$ are the state and action spaces, $P: \mathcal{S} \times \mathcal{A} \rightarrow \mathcal{S}$ is the transition function, $R: \mathcal{S} \times \mathcal{A} \rightarrow \mathbb{R}$ is the reward function, $H$ is the horizon, and $\gamma$ is the discount factor. In agentic RL, both states and actions are represented in natural language. At step $t$, the state $s_t$ typically contains the task specification, interaction history, and additional action instructions. An episode produces a trajectory $\tau = \{(s_1, a_1, r_1), (s_2, a_2, r_2), \cdots, (s_T, a_T, r_T)\}$, where $T \leq H$ is the episode length. Because the environment usually provides only a terminal outcome reward, we set $\gamma = 1$ and denote the trajectory-level reward by $R(\tau)$.

\noindent \textbf{LLM-driven agent policy.} The agent policy $\pi_\theta$, parameterized by $\theta$, is instantiated by an LLM and can be described at two equivalent levels. At the action level, the policy maps the current state $s_t$ to an action executed in the environment, i.e., $a_t \sim \pi_\theta(\cdot \mid s_t)$. At the token level, each action is generated auto-regressively as a token sequence $a_t = \{a_{t,k}\}_{k=1}^{K_t}$, where $a_{t,k} \sim \pi_\theta(\cdot \mid s_t, a_{t,<k})$. Rewards are defined at the action level and broadcast to all tokens in the action. The RL objective is:
\begin{equation}
    J_{\text{RL}}(\theta) = \mathbb{E}_{\tau \sim \pi_{\theta}} \left[ R(\tau) \right].
\end{equation}
We further assume a non-cumulative setting~\citep{feng2025group}, in which each step is generated from an individual prompt constructed from the current state rather than from the full interaction history. This design helps stabilize training by keeping the effective context length under control.

\noindent \textbf{Group-based reinforcement learning.} Classical actor-critic methods, such as PPO~\citep{schulman2017proximal,ouyang2022training}, jointly optimize an actor and a critic. Although effective, they introduce additional computation and memory cost. In contrast, group-based RL methods represented by GRPO~\citep{shao2024deepseekmath} remove the explicit critic and derive the learning signal from relative reward comparisons within a sampled group. This formulation extends naturally to agentic tasks by assigning the trajectory-level reward to all actions in the episode. To be specific, for each task, GRPO samples a group of $G$ trajectories $\{\tau^i\}_{i=1}^G$ from the old policy $\pi_{\theta_{\text{old}}}$. Each trajectory receives a scalar reward $r^i = R(\tau^i)$, which is then normalized within the group to produce the corresponding advantage:
\begin{equation}
    A^i = \frac{r^i - \text{mean}(\{r^i\}_{i=1}^G)}{\text{std}(\{r^i\}_{i=1}^G)}.
    \label{eq:advantage_computation}
\end{equation}
For simplicity, we define the generic clipped surrogate as:
\begin{equation}
    \mathcal{L}(x, A; \theta) = \frac{1}{|x|} \sum_{k=1}^{|x|} \min \left( \rho_{x,k}(\theta) A, \text{clip}\left(\rho_{x,k}(\theta), 1-\epsilon, 1+\epsilon\right) A \right),
    \label{eq:clipped_surrogate}
\end{equation}
where $x$ is an arbitrary response, $|x|$ denotes the number of tokens in $x$, $A$ denotes the corresponding advantage, and $\rho_{x,k}(\theta) = \frac{\pi_\theta(x_k \mid s_t, x_{<k})}{\pi_{\theta_{\text{old}}}(x_k \mid s_t, x_{<k})}$ defines the importance sampling ratio for token $x_k$. Then GRPO optimizes the following objective built upon the generic clipped surrogate:
\begin{equation}
    \mathcal{J}_{\text{GRPO}}(\theta) = \mathbb{E}_{\{\tau^i\}_{i=1}^G \sim \pi_{\theta}} \left[ \frac{1}{G} \sum_{i=1}^G \frac{1}{|\tau^i|} \sum_{t=1}^{|\tau^i|} \mathcal{L}(a_t^i, A^i; \theta) - \beta D_{\text{KL}}\left( \pi_\theta \| \pi_{\text{ref}} \right) \right],
    \label{eq:grpo_objective}
\end{equation}
where $|\tau^i|$ is the number of steps in trajectory $\tau^i$. By forgoing the explicit critic, group-based RL achieves significant gains in computational and memory efficiency while maintaining scalability, making it particularly well-suited for training LLM-based agents.

\section{Method}
\label{sec:method}

Motivated by how humans solve complex interactive tasks in real-world scenarios, we propose \textbf{Stra}tegic \textbf{T}rajectory \textbf{A}bstraction (StraTA), a framework that introduces explicit strategy guidance into agentic RL. Instead of generating actions purely reactively from the current state, StraTA first produces a compact natural-language strategy at the beginning of each episode and conditions all subsequent actions on it. This design decomposes learning into two simpler objectives: (1) generating feasible strategies, and (2) executing actions that consistently follow them.

\begin{figure}[t]
    \centering
    \includegraphics[width=1.0\textwidth]{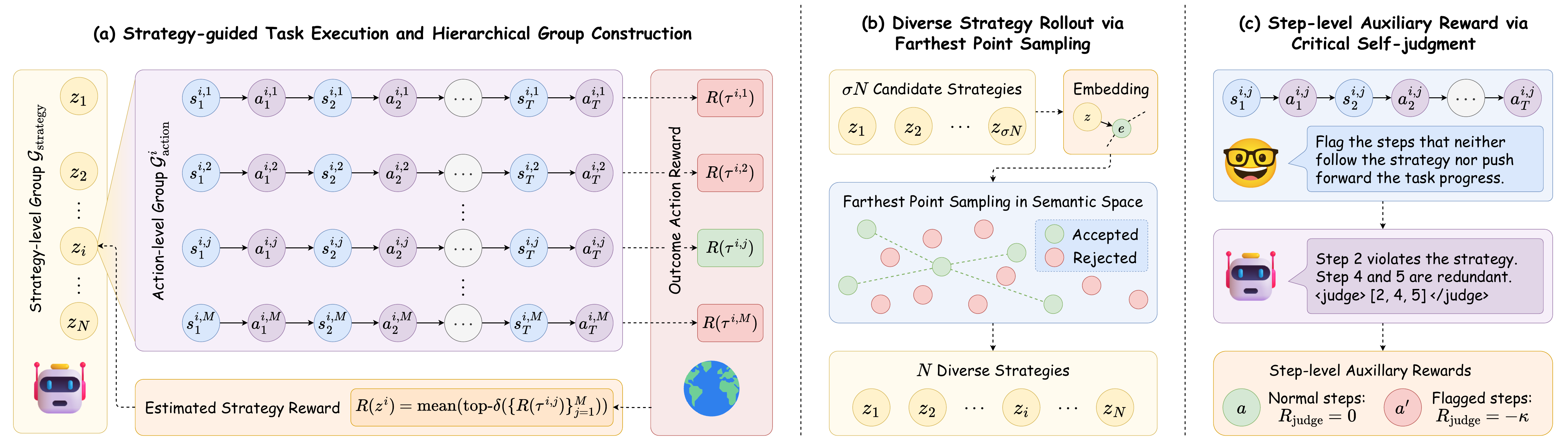}
    \caption{An overview of the StraTA framework. (a) For a given task, StraTA generates multiple strategies and performs multiple rollouts for each strategy, forming hierarchical groups for strategy-level and action-level optimization. (b) Diverse strategy rollout uses farthest point sampling to select semantically distinct strategies, improving exploration of the strategy space. (c) Critical self-judgment identifies action steps that neither follow the strategy nor advance task progress, and assigns a step-level auxiliary penalty to achieve more fine-grained credit assignment.}
    \label{fig:method}
\end{figure}

In this section, we will describe StraTA in four parts as shown in Figure~\ref{fig:method}. First, we will present the strategy-guided task execution process. Then, we will explain how StraTA samples trajectories and constructs hierarchical groups for GRPO-style training. Finally, we will introduce two techniques that respectively support the two learning objectives mentioned above: (1) diverse strategy rollout via farthest point sampling, and (2) step-level auxiliary reward via critical self-judgment.

\subsection{Strategy-guided Task Execution}
\label{subsec:strategy_guided_task_execution}

In typical agentic RL, the agent acts reactively: at each step $t$, it generates the current action $a_t$ directly from the current state $s_t$, i.e., $a_t \sim \pi_\theta(\cdot \mid s_t)$. Under this formulation, the trajectory is determined entirely by step-by-step decisions, without explicit trajectory-level guidance. The policy therefore has to infer both the long-term course of action and the immediate next action at the same time, which often leads to short-sighted behavior, unnecessary backtracking, and inconsistent decisions.

StraTA augments this process by inserting a strategy generation step before interaction begins. Given the initial state $s_1$, which contains the task description, the agent first samples a global strategy:
\begin{equation}
    z \sim \pi_\theta(\cdot \mid s_1),
\end{equation}
where $z$ is a compact natural-language plan that summarizes the anticipated course of action for the entire episode. The strategy is generated once at the beginning of the episode and kept fixed thereafter. Conditioned on $z$, action generation is no longer purely reactive. At each step $t$, the agent generates its action $a_t$ based on both the global strategy $z$ and the current state $s_t$:
\begin{equation}
    a_t \sim \pi_\theta(\cdot \mid z, s_t),
\end{equation}
where $z$ serves as a trajectory-level signal that is prepended to the state at every step, so each action is generated from both the global strategy and the local context. The resulting trajectory becomes
\begin{equation}
    \tau = \{z, (s_1, a_1, r_1), \cdots, (s_T, a_T, r_T)\}.
\end{equation}
This reformulation decomposes task solving into two sub-goals: (1) generating high-quality strategies from the initial task specification, and (2) executing actions that consistently follow the generated strategy throughout the episode. By separating these two roles, StraTA turns the original fully reactive decision process into a more structured one. Long-horizon behavior can now be coordinated through an explicit strategy instead of being inferred implicitly from local action choices alone. We will next describe how to train this strategy-guided policy with a GRPO-style objective.

\subsection{Hierarchical Group Construction}
\label{subsec:hierarchical_group_construction}

To jointly optimize strategy generation and action taking, StraTA organizes rollouts into a two-level hierarchy on top of GRPO-style training. Figure~\ref{fig:method}a intuitively illustrates the sampling process. For each task, the agent first samples $N$ strategies $\{z^i\}_{i=1}^N$ and then performs $M$ independent rollouts $\{\tau^{i,j}\}_{j=1}^M$ under each strategy $z^i$, which yields $N \times M$ trajectories in total. This construction induces one strategy-level group and $N$ action-level groups:
\begin{equation}
    \mathcal{G}_{\text{strategy}} = \{z^i \mid 1\leq i\leq N\},\quad \mathcal{G}_{\text{action}}^i = \{a_t^{i,j} \mid 1\leq j\leq M, 1\leq t\leq |\tau^{i,j}| \}.
\end{equation}
The strategy-level group compares multiple strategies, allowing the model to identify which high-level plans are more promising. Each action-level group compares multiple rollouts under the same strategy, revealing which trajectories are more likely to solve the task while adhering to that strategy.

At the action level, each trajectory $\tau^{i,j}$ receives an outcome reward $R(\tau^{i,j})$ at the end of the episode. Because we adopt a non-cumulative setting, this trajectory-level reward should be broadcast to all action steps $\{a_t^{i,j}\}_{t=1}^{|\tau^{i,j}|}$ within the same episode:
\begin{equation}
    R(a_t^{i,j}) = R(\tau^{i,j}),
\end{equation}
so that every action step can receive a learning signal even though the environment only returns reward at episode termination.

At the strategy level, each strategy $z^i$ is assigned a reward based on the rollouts $\{\tau^{i,j}\}_{j=1}^M$ generated under it. Instead of averaging over all rollouts equally, StraTA uses the mean of the top-$\delta$ fraction of outcome rewards, where $\delta$ controls the fraction of rewards to be considered:
\begin{equation}
    R(z^i) = \text{mean}\left( \text{top-}\delta \left( \{R(\tau^{i,j})\}_{j=1}^M \right) \right),
\end{equation}
which emphasizes the best performance achievable under a strategy and reduces the influence of low-quality rollouts caused by early-stage action noise. As a result, the reward more faithfully reflects strategy quality rather than imperfections in action execution.

Both the action-level and strategy-level rewards are augmented with two auxiliary terms. GRPO-style training often encourages long responses. To keep generations within the context budget, we follow prior work~\citep{yeo2025demystifying,liu2025understanding,yu2025dapo} and introduce a soft length penalty:
\begin{equation}
    R_{\text{length}}(x) = \left\{ \begin{aligned}
        &0, \quad &&|x| \leq \lambda L_{\text{total}} \\
        &-\frac{1}{1-\lambda} \left( \frac{|x|}{L_{\text{total}}} - \lambda \right), &&\lambda L_{\text{total}} < |x| \leq L_{\text{total}} \\
        &-1, &&L_{\text{total}} < |x|  \\
    \end{aligned} \right.,
\end{equation}
where $x$ can be either a strategy or an action, $|x|$ is the number of tokens in $x$, $L_{\text{total}}$ is the response length limit, and $\lambda$ controls when the penalty begins. This term softly discourages overlong generations and helps stabilize training. In addition, to encourage responses that can be executed by the environment, we introduce a hard format penalty:
\begin{equation}
    R_{\text{format}}(x) = \left\{ \begin{aligned}
        &0, \quad &&x\text{ is correct in format} \\
        &-1, &&x\text{ is incorrect in format}  \\
    \end{aligned} \right.,
\end{equation}
which penalizes only unrecognized responses and further stabilizes training. The final reward is
\begin{equation}
    \hat{R}(x) = \text{clip} \left(R(x) + R_{\text{length}}(x) + R_{\text{format}}(x), -1, 1 \right),
    \label{eq:reward_formulation}
\end{equation}
which clips the total reward to $[-1, 1]$ and applies to both strategy and action steps.

All resulting groups are compatible with the GRPO-style training introduced in Section~\ref{sec:preliminaries}. Accordingly, advantage computation in Equation~\ref{eq:advantage_computation} is applied unchanged to both the strategy-level group $\mathcal{G}_{\text{strategy}}$ and each action-level group $\mathcal{G}_{\text{action}}^i$. The overall StraTA training objective is:
\begin{equation}
    \begin{aligned}
        \mathcal{J}_{\text{StraTA}}(\theta) = \mathbb{E}_{\mathcal{G}_{\text{strategy}}, \{\mathcal{G}_{\text{action}}^i\}_{i=1}^N \sim \pi_{\theta}} &\left[ \sum_{z^i \in \mathcal{G}_{\text{strategy}}} \mathcal{L}(z^i, A(z^i); \theta) \right. \\
        &\left. + \sum_{i=1}^N \sum_{a_t^{i,j} \in \mathcal{G}_{\text{action}}^i} \mathcal{L}(a_t^{i,j}, A(a_t^{i,j}); \theta) - \beta D_{\text{KL}}\left( \pi_\theta \| \pi_{\text{ref}} \right) \right],
    \end{aligned}
\end{equation}
where $A(z^i)$ and $A(a_t^{i,j})$ denote the computed advantages for $z^i$ and $a_t^{i,j}$, respectively.

\subsection{Diverse Strategy Rollout via Farthest Point Sampling}
\label{subsec:diverse_strategy_rollout}

Prior work~\citep{cui2025entropy,chen2025minimax,yu2025dapo} has shown that rollout diversity is crucial for agentic RL. However, even with elevated sampling temperature, the model produces highly similar strategies within the same group, which limits exploration of the strategy space.

As shown in Figure~\ref{fig:method}b, to encourage diverse strategies, StraTA first oversamples $\sigma \times N$ candidate strategies and then applies farthest point sampling to select $N$ strategies that are maximally different from one another, where $\sigma$ controls the oversampling ratio. To quantify semantic diversity, each candidate strategy $z^i$ is encoded as a normalized vector $e^i = \phi(z^i)$ using a pre-trained embedding model $\phi(\cdot)$~\citep{reimers2019sentence}, so the inner product ${e^i}^{\top} e^j$ is the cosine similarity.

The selection proceeds greedily. We initialize the selected set $\mathcal{Z}$ with the candidate $z_{\text{center}}$ that is closest to the centroid, yielding a representative starting point. At each subsequent round, we add the candidate with the smallest maximum similarity to the already selected set $\mathcal{Z}$:
\begin{equation}
z_{\text{select}} = \arg \min_{z^i \notin \mathcal{Z}} \max_{z^j \in \mathcal{Z}} {e^i}^{\top} e^j.
\end{equation}
This process repeats until $N$ strategies have been selected, producing a diverse set of strategy rollouts. As a result, strategy-level comparison in GRPO becomes more informative because the sampled strategies provide meaningful contrast rather than minor variations. It also exposes the agent to a broader range of strategic behaviors, encouraging more effective exploration.

\subsection{Step-level Auxiliary Reward via Critical Self-judgment}
\label{subsec:step_level_auxiliary_reward}

Another key challenge is credit assignment. Sparse outcome rewards make it difficult to identify which actions truly contribute to task success. Moreover, optimizing only for final success may favor opportunistic actions that work in hindsight but do not faithfully follow the intended strategy.

As shown in Figure~\ref{fig:method}c, to address this issue, StraTA introduces a self-judgment mechanism inspired by LLM-as-a-judge techniques~\citep{zheng2023judging,gu2024survey,li2025generation} to achieve fine-grained credit assignment. After a rollout is completed, the agent retrospectively evaluates the trajectory against its guiding strategy. Concretely, given an episode $\tau^{i,j}$ under strategy $z^i$, the agent is instructed to flag all action steps that neither contribute to the final success nor follow the global strategy:
\begin{equation}
    \mathcal{I}^{i,j} \sim \pi_\theta(\cdot \mid \mathcal{P}, \tau^{i,j}),
\end{equation}
where $\mathcal{P}$ is the self-judgment prompt and $\mathcal{I}^{i,j}$ is the set of flagged action-step indices. The self-judgment is performed by the agent itself rather than by a separate judge model, and these judging steps do not receive reward. The resulting step-level auxiliary reward for each action $a_t^{i,j}$ is
\begin{equation}
    R_{\text{judge}}(a_t^{i,j}) = -\kappa \cdot \mathds{1}[t \in \mathcal{I}^{i,j}],
\end{equation}
where $\kappa$ controls the weight of the self-judgment penalty. This term is then additionally added to the action reward, replacing Equation~\ref{eq:reward_formulation} for action steps:
\begin{equation}
    \hat{R}(a_t^{i,j}) = \text{clip} \left(R(a_t^{i,j}) + R_{\text{length}}(a_t^{i,j}) + R_{\text{format}}(a_t^{i,j}) + R_{\text{judge}}(a_t^{i,j}), -1, 1\right).
\end{equation}
Therefore, the final action reward combines coarse outcome supervision with fine-grained self-judgment. This design encourages the policy to optimize not only for successful rollouts, but also for intermediate actions that are genuinely aligned with the intended strategy.

\section{Experiments}
\label{sec:experiments}

\begin{table}[t]
    \caption{Performance on ALFWorld and WebShop. Results from open-source models are averaged over 3 random seeds. For ALFWorld, we report the success rate of each subtask together with the overall success rate. For WebShop, we report the overall task score and overall success rate. The \hlfirst{best} and \hlsecond{second best} results are highlighted. $^\dagger$The results are from the original paper~\citep{feng2025group}.}
    \label{tab:alfworld_and_webshop_results}
    \renewcommand{\arraystretch}{1.0}
    \centering \footnotesize \setlength{\tabcolsep}{3pt}
    \begin{tabular}{lccccccc|cc}
        \toprule
            \multicolumn{1}{c}{\multirow{2}{*}{\textbf{Method}}} & \multicolumn{7}{c|}{\textbf{ALFWorld}} & \multicolumn{2}{c}{\textbf{WebShop}} \\
            & Pick & Look & Clean & Heat & Cool & Pick2 & All & Score & Succ \\
        \midrule
            \multicolumn{10}{l}{\textit{Closed-Source Model}} \\
                \quad GPT-5.1 & $\text{80.0}$ & $\text{53.8}$ & $\text{22.2}$ & $\text{62.5}$ & $\text{44.0}$ & $\text{66.7}$ & $\text{55.7}$ & $\text{31.2}$ & $\text{22.2}$ \\
                \quad Claude-4-Sonnet & $\text{80.0}$ & $\text{76.9}$ & $\text{63.0}$ & $\text{75.0}$ & $\text{72.0}$ & $\text{70.8}$ & $\text{72.9}$ & $\text{23.7}$ & $\text{18.4}$ \\
                \quad Gemini-2.5-Flash & $\text{71.4}$ & $\text{30.8}$ & $\text{40.7}$ & $\text{18.8}$ & $\text{20.0}$ & $\text{25.0}$ & $\text{38.6}$ & $\text{4.7}$ & $\text{4.4}$ \\
        \midrule
            \multicolumn{10}{l}{\textit{Qwen2.5-1.5B-Instruct}} \\
                \quad Vanilla & $\text{5.7}_{\pm \text{0.0}}$ & $\text{25.6}_{\pm \text{4.4}}$ & $\text{1.2}_{\pm \text{2.1}}$ & $\text{0.0}_{\pm \text{0.0}}$ & $\text{0.0}_{\pm \text{0.0}}$ & $\text{2.8}_{\pm \text{2.4}}$ & $\text{4.5}_{\pm \text{0.4}}$ & $\text{20.2}_{\pm \text{0.7}}$ & $\text{2.9}_{\pm \text{0.8}}$ \\
                \quad ReAct & $\text{30.5}_{\pm \text{1.3}}$ & $\text{12.8}_{\pm \text{9.6}}$ & $\text{7.4}_{\pm \text{3.0}}$ & $\text{18.8}_{\pm \text{8.8}}$ & $\text{17.3}_{\pm \text{1.9}}$ & $\text{12.5}_{\pm \text{6.8}}$ & $\text{17.6}_{\pm \text{2.0}}$ & $\text{16.2}_{\pm \text{0.3}}$ & $\text{9.1}_{\pm \text{0.4}}$ \\
                \quad PPO & $\text{86.7}_{\pm \text{4.4}}$ & $\text{61.5}_{\pm \text{7.7}}$ & \hlsecond{$\text{93.8}_{\pm \text{2.1}}$} & \hlfirst{$\text{97.9}_{\pm \text{3.6}}$} & \hlsecond{$\text{84.0}_{\pm \text{0.0}}$} & \hlsecond{$\text{80.6}_{\pm \text{2.4}}$} & $\text{85.5}_{\pm \text{2.2}}$ & $\text{78.2}_{\pm \text{0.4}}$ & $\text{52.8}_{\pm \text{0.2}}$ \\
                \quad RLOO$^\dagger$ & $\text{88.3}_{\pm \text{3.0}}$ & $\text{52.8}_{\pm \text{8.6}}$ & $\text{71.0}_{\pm \text{5.9}}$ & $\text{62.8}_{\pm \text{8.7}}$ & $\text{66.4}_{\pm \text{5.5}}$ & $\text{56.9}_{\pm \text{4.7}}$ & $\text{69.7}_{\pm \text{2.5}}$ & $\text{73.9}_{\pm \text{5.6}}$ & $\text{52.1}_{\pm \text{6.7}}$ \\
                \quad GRPO & $\text{80.0}_{\pm \text{0.0}}$ & $\text{61.5}_{\pm \text{0.0}}$ & $\text{81.5}_{\pm \text{3.7}}$ & \hlsecond{$\text{95.8}_{\pm \text{7.2}}$} & $\text{62.7}_{\pm \text{4.6}}$ & $\text{69.4}_{\pm \text{4.8}}$ & $\text{75.5}_{\pm \text{2.9}}$ & $\text{80.0}_{\pm \text{0.5}}$ & $\text{62.2}_{\pm \text{1.2}}$ \\
                \quad GiGPO$^\dagger$ & \hlfirst{$\text{94.4}_{\pm \text{5.9}}$} & \hlsecond{$\text{67.5}_{\pm \text{4.6}}$} & \hlfirst{$\text{94.8}_{\pm \text{3.8}}$} & $\text{94.4}_{\pm \text{7.8}}$ & $\text{79.8}_{\pm \text{4.7}}$ & $\text{76.4}_{\pm \text{5.4}}$ & \hlsecond{$\text{86.7}_{\pm \text{1.7}}$} & \hlsecond{$\text{83.1}_{\pm \text{1.6}}$} & \hlsecond{$\text{65.0}_{\pm \text{3.2}}$} \\
                \quad StraTA (Ours) & \hlsecond{$\text{89.5}_{\pm \text{1.6}}$} & \hlfirst{$\text{87.2}_{\pm \text{4.4}}$} & $\text{92.6}_{\pm \text{0.0}}$ & $\text{93.8}_{\pm \text{0.0}}$ & \hlfirst{$\text{90.7}_{\pm \text{4.6}}$} & \hlfirst{$\text{90.3}_{\pm \text{2.4}}$} & \hlfirst{$\text{90.7}_{\pm \text{1.9}}$} & \hlfirst{$\text{91.1}_{\pm \text{0.3}}$} & \hlfirst{$\text{82.5}_{\pm \text{0.6}}$} \\
        \midrule
            \multicolumn{10}{l}{\textit{Qwen2.5-7B-Instruct}} \\
                \quad Vanilla & $\text{50.5}_{\pm \text{9.1}}$ & $\text{43.6}_{\pm \text{4.4}}$ & $\text{27.2}_{\pm \text{5.7}}$ & $\text{14.6}_{\pm \text{7.2}}$ & $\text{16.0}_{\pm \text{4.0}}$ & $\text{5.6}_{\pm \text{4.8}}$ & $\text{27.4}_{\pm \text{1.5}}$ & $\text{31.2}_{\pm \text{1.0}}$ & $\text{5.3}_{\pm \text{1.5}}$ \\
                \quad ReAct & $\text{56.2}_{\pm \text{3.6}}$ & $\text{43.6}_{\pm \text{3.6}}$ & $\text{49.4}_{\pm \text{4.6}}$ & $\text{43.8}_{\pm \text{5.1}}$ & $\text{37.3}_{\pm \text{6.8}}$ & $\text{18.1}_{\pm \text{5.2}}$ & $\text{42.4}_{\pm \text{1.7}}$ & $\text{35.7}_{\pm \text{0.2}}$ & $\text{27.3}_{\pm \text{0.2}}$ \\
                \quad PPO & $\text{96.2}_{\pm \text{1.7}}$ & $\text{71.8}_{\pm \text{8.9}}$ & $\text{96.3}_{\pm \text{0.0}}$ & \hlsecond{$\text{91.7}_{\pm \text{3.6}}$} & $\text{70.7}_{\pm \text{6.1}}$ & \hlfirst{$\text{87.5}_{\pm \text{0.0}}$} & $\text{87.4}_{\pm \text{1.1}}$ & $\text{82.2}_{\pm \text{0.2}}$ & $\text{68.9}_{\pm \text{1.0}}$ \\
                \quad RLOO$^\dagger$ & $\text{87.6}_{\pm \text{4.3}}$ & $\text{78.2}_{\pm \text{8.3}}$ & $\text{87.3}_{\pm \text{5.8}}$ & $\text{81.3}_{\pm \text{7.6}}$ & $\text{71.9}_{\pm \text{5.2}}$ & $\text{48.9}_{\pm \text{8.4}}$ & $\text{75.5}_{\pm \text{4.6}}$ & $\text{80.3}_{\pm \text{3.2}}$ & $\text{65.7}_{\pm \text{4.0}}$ \\
                \quad GRPO & \hlfirst{$\text{98.1}_{\pm \text{2.7}}$} & $\text{71.8}_{\pm \text{9.6}}$ & $\text{85.2}_{\pm \text{3.0}}$ & $\text{58.3}_{\pm \text{7.8}}$ & $\text{73.3}_{\pm \text{5.0}}$ & $\text{69.4}_{\pm \text{3.9}}$ & $\text{79.3}_{\pm \text{1.5}}$ & $\text{84.1}_{\pm \text{0.7}}$ & $\text{71.5}_{\pm \text{0.7}}$ \\
                \quad GiGPO$^\dagger$ & \hlsecond{$\text{97.7}_{\pm \text{1.6}}$} & \hlsecond{$\text{82.7}_{\pm \text{7.9}}$} & \hlfirst{$\text{98.8}_{\pm \text{1.6}}$} & $\text{83.7}_{\pm \text{7.2}}$ & \hlsecond{$\text{89.3}_{\pm \text{8.2}}$} & $\text{79.2}_{\pm \text{6.6}}$ & \hlsecond{$\text{90.8}_{\pm \text{1.3}}$} & \hlsecond{$\text{84.4}_{\pm \text{2.9}}$} & \hlsecond{$\text{72.8}_{\pm \text{3.2}}$} \\
                \quad StraTA (Ours) & $\text{94.3}_{\pm \text{0.0}}$ & \hlfirst{$\text{92.3}_{\pm \text{0.0}}$} & \hlsecond{$\text{97.5}_{\pm \text{2.1}}$} & \hlfirst{$\text{100.0}_{\pm \text{0.0}}$} & \hlfirst{$\text{93.3}_{\pm \text{4.6}}$} & \hlsecond{$\text{81.9}_{\pm \text{2.4}}$} & \hlfirst{$\text{93.1}_{\pm \text{0.8}}$} & \hlfirst{$\text{91.2}_{\pm \text{0.3}}$} & \hlfirst{$\text{84.2}_{\pm \text{0.3}}$} \\
        \bottomrule
    \end{tabular}
\end{table}

\begin{table}[t]
    \caption{Performance on SciWorld. Results from open-source models are averaged over 3 random seeds. We report the score of each task category together with the overall score. The \hlfirst{best} and \hlsecond{second best} results are highlighted. $^\dagger$The results are from the original paper~\citep{xi2025agentgymrl}.}
    \label{tab:sciworld_results}
    \renewcommand{\arraystretch}{1.0}
    \centering \footnotesize \setlength{\tabcolsep}{3pt}
    \begin{tabular}{llcccccc}
        \toprule
            \multicolumn{1}{c}{\multirow{2}{*}{\textbf{Method}}} & \multicolumn{1}{c}{\multirow{2}{*}{\textbf{Category}}} & \multicolumn{6}{c}{\textbf{SciWorld}} \\
            & & Measure & Test-Cond & Find & Chem-Mix & Lifespan & All \\
        \midrule
            \multicolumn{8}{l}{\textit{Closed-Source Model}} \\
                \quad GPT-5.1 & Prompting-Based & $\text{32.1}$ & $\text{46.5}$ & $\text{41.8}$ & $\text{69.6}$ & $\text{38.3}$ & $\text{43.0}$ \\
                \quad Claude-4-Sonnet & Prompting-Based & $\text{45.4}$ & $\text{54.7}$ & $\text{60.3}$ & $\text{45.6}$ & $\text{90.0}$ & $\text{57.4}$ \\
                \quad Gemini-2.5-Flash & Prompting-Based & $\text{40.7}$ & $\text{51.3}$ & $\text{53.7}$ & $\text{33.0}$ & $\text{50.0}$ & $\text{49.6}$ \\
        \midrule
            \multicolumn{8}{l}{\textit{Qwen2.5-7B-Instruct}} \\
                \quad AgentGym-RL$^\dagger$ & Training-Based & $\text{24.5}$ & \hlfirst{$\text{59.0}$} & $\text{65.9}$ & $\text{0.0}$ & $\text{66.7}$ & $\text{50.5}$ \\
                \quad ScalingInter$^\dagger$ & Training-Based & \hlsecond{$\text{34.0}$} & $\text{55.4}$ & \hlfirst{$\text{88.6}$} & $\text{0.0}$ & $\text{73.3}$ & \hlsecond{$\text{57.0}$} \\
                \quad Vanilla & Prompting-Based & $\text{2.1}_{\pm \text{1.5}}$ & $\text{12.3}_{\pm \text{1.3}}$ & $\text{21.2}_{\pm \text{1.6}}$ & $\text{9.6}_{\pm \text{3.3}}$ & $\text{25.0}_{\pm \text{2.4}}$ & $\text{13.9}_{\pm \text{0.8}}$ \\
                \quad ReAct & Prompting-Based & $\text{12.2}_{\pm \text{2.1}}$ & $\text{14.4}_{\pm \text{1.2}}$ & $\text{33.6}_{\pm \text{2.7}}$ & $\text{9.6}_{\pm \text{2.8}}$ & $\text{28.9}_{\pm \text{3.4}}$ & $\text{20.0}_{\pm \text{0.3}}$ \\
                \quad PPO & Training-Based & $\text{1.8}_{\pm \text{1.3}}$ & \hlsecond{$\text{57.5}_{\pm \text{0.2}}$} & $\text{60.9}_{\pm \text{0.8}}$ & $\text{6.9}_{\pm \text{1.9}}$ & \hlsecond{$\text{97.8}_{\pm \text{3.1}}$} & $\text{51.4}_{\pm \text{0.3}}$ \\
                \quad GRPO & Training-Based & $\text{31.2}_{\pm \text{4.8}}$ & $\text{49.8}_{\pm \text{2.9}}$ & $\text{32.7}_{\pm \text{3.6}}$ & \hlsecond{$\text{14.7}_{\pm \text{5.4}}$} & $\text{53.3}_{\pm \text{5.0}}$ & $\text{41.8}_{\pm \text{2.2}}$ \\
                \quad StraTA (Ours) & Training-Based & \hlfirst{$\text{55.7}_{\pm \text{3.2}}$} & $\text{54.4}_{\pm \text{0.7}}$ & \hlsecond{$\text{78.6}_{\pm \text{1.1}}$} & \hlfirst{$\text{16.3}_{\pm \text{1.2}}$} & \hlfirst{$\text{100.0}_{\pm \text{0.0}}$} & \hlfirst{$\text{63.5}_{\pm \text{0.7}}$} \\
        \bottomrule
    \end{tabular}
\end{table}

\subsection{Experimental Setup}
\label{subsec:experimental_setup}

\noindent \textbf{Benchmarks.} We evaluate StraTA on three representative benchmarks: ALFWorld~\citep{shridhar2021alfworld}, WebShop~\citep{yao2022webshop}, and SciWorld~\citep{wang2022scienceworld}, covering diverse scenarios including embodied manipulation, web navigation, and scientific experimentation. To standardize evaluation across environments, we serve all benchmarks through AgentGym~\citep{xi2025agentgym}, a unified agentic RL codebase. Following GiGPO~\citep{feng2025group} and AgentGym-RL~\citep{xi2025agentgymrl}, we report success rate on ALFWorld with a detailed per-subtask breakdown, both task score and success rate on WebShop, and task score on SciWorld.

\noindent \textbf{Baselines.} We compare StraTA against diverse baselines spanning prompting-based and training-based approaches. For prompting-based methods, we include the frontier closed-source models GPT-5.1~\citep{openai2025gpt}, Claude-4-Sonnet~\citep{anthropic2025introducing}, and Gemini-2.5-Flash~\citep{comanici2025gemini}, together with the open-source Qwen2.5 series. In addition to vanilla prompting, which generates actions directly without explicit reasoning, we include ReAct~\citep{yao2023react} as a stronger baseline that interleaves reasoning and action generation at each step. For training-based approaches, we consider standard RL methods: PPO~\citep{schulman2017proximal}, RLOO~\citep{ahmadian2024back}, GRPO~\citep{shao2024deepseekmath}, and GiGPO~\citep{feng2025group}. PPO optimizes the policy using a critic network, whereas RLOO replaces the critic with leave-one-out reward estimation. GRPO derives learning signals from relative reward comparisons within groups. GiGPO further improves long-horizon performance by introducing anchor states for credit assignment. For SciWorld, we additionally include AgentGym-RL-7B and ScalingInter-7B~\citep{xi2025agentgymrl} as strong RL-trained baselines.

\noindent \textbf{Training details.} Our training procedure primarily follows GiGPO~\citep{feng2025group}, with additional specifics aligned with AgentGym-RL~\citep{xi2025agentgymrl}. For our backbone models, we utilize Qwen2.5-1.5B-Instruct and Qwen2.5-7B-Instruct as representative architectures of different scales~\citep{yang2024qwen}. StraTA is trained for 150 steps with a batch size of 16, where each task samples 4 strategies and performs 8 rollouts per strategy, resulting in 32 rollouts per task. The oversampling ratio $\sigma$ is set to 8, the aggregation ratio $\delta$ to 0.5, the length penalty threshold $\lambda$ to 0.5, and the self-judgment reward weight $\kappa$ to 0.1. For fair comparison, we prioritize aligning the amount of training tasks, and thus adopt a batch size of 16 for all methods. For the GRPO-style methods we reproduce, we further set the number of rollouts per task to 32 so that their computational budget matches that of StraTA. All experiments are run on a single node equipped with 8 NVIDIA H100 GPUs. Further implementation details and complete parameter specifications are available in Appendix~\ref{appsubsec:training_parameters}.

\subsection{Main Results}
\label{subsec:main_results}

\noindent \textbf{Performance on ALFWorld and WebShop.} Table~\ref{tab:alfworld_and_webshop_results} reports the results on ALFWorld and WebShop. We highlight three key observations. (1) Models without RL training fail to reliably solve these tasks, suggesting that specialized training is necessary for long-horizon agentic problems. Frontier closed-source models only achieve the best success rates reaching 72.9\% on ALFWorld and 22.2\% on WebShop. (2) RL-based methods consistently improve performance and clearly outperform frontier closed-source models. PPO achieves success rates of 87.4\% and 68.9\% on ALFWorld and WebShop, respectively, while GRPO and GiGPO perform even better in some settings. (3) StraTA delivers the strongest overall performance. At the 1.5B scale, StraTA reaches success rates of 90.7\% on ALFWorld and 82.5\% on WebShop. At the 7B scale, StraTA further improves to 93.1\% and 84.2\%. As shown in Figure~\ref{fig:teaser}c, the training curves on WebShop intuitively show the efficiency of StraTA.

\noindent \textbf{Performance on SciWorld.} Table~\ref{tab:sciworld_results} reports the results on SciWorld. We emphasize three observations. (1) Frontier closed-source models substantially outperform lightweight open-source models, which reflects the greater difficulty of SciWorld. (2) RL training can raise open-source models to the level of frontier models: ScalingInter achieves an overall score of 57.0\%, and our trained PPO and GRPO approximately reproduce this result. (3) StraTA again achieves state-of-the-art performance, reaching an overall score of 63.5\% and even attaining a perfect 100.0\% score on the Lifespan subset.

\subsection{Ablation Studies}
\label{subsec:ablation_studies}

\begin{table}[ht]
    \caption{Performance comparison of StraTA with different variants. \textit{Vanilla} means no trick is applied. \textit{Diverse} represents the version with diverse strategy rollout via farthest point sampling. \textit{Judgment} represents the version with step-level auxiliary reward via critical self-judgment.}
    \label{tab:ablation_components}
    \renewcommand{\arraystretch}{1.0}
    \centering \footnotesize \setlength{\tabcolsep}{3pt}
    \begin{tabular}{lccccccc|cc}
        \toprule
            \multicolumn{1}{c}{\multirow{2}{*}{\textbf{Variant}}} & \multicolumn{7}{c|}{\textbf{ALFWorld}} & \multicolumn{2}{c}{\textbf{WebShop}} \\
            & Pick & Look & Clean & Heat & Cool & Pick2 & All & Score & Succ \\
        \midrule
            \multicolumn{10}{l}{\textit{Qwen2.5-3B-Instruct}} \\
                \quad Vanilla & $\text{87.6}_{\pm \text{2.7}}$ & $\text{69.2}_{\pm \text{6.3}}$ & $\text{88.9}_{\pm \text{0.0}}$ & $\text{66.7}_{\pm \text{5.9}}$ & $\text{70.7}_{\pm \text{1.9}}$ & $\text{77.8}_{\pm \text{2.0}}$ & $\text{79.0}_{\pm \text{2.2}}$ & $\text{76.0}_{\pm \text{0.2}}$ & $\text{64.0}_{\pm \text{0.4}}$ \\
                \quad Diverse & \hlsecond{$\text{92.4}_{\pm \text{1.3}}$} & \hlfirst{$\text{84.6}_{\pm \text{0.0}}$} & \hlfirst{$\text{96.3}_{\pm \text{0.0}}$} & \hlsecond{$\text{85.4}_{\pm \text{2.9}}$} & \hlfirst{$\text{78.7}_{\pm \text{1.9}}$} & \hlsecond{$\text{84.7}_{\pm \text{2.0}}$}
                & \hlsecond{$\text{87.9}_{\pm \text{0.6}}$} & $\text{75.9}_{\pm \text{0.6}}$ & $\text{64.6}_{\pm \text{0.2}}$ \\
                \quad Judgment & $\text{88.6}_{\pm \text{4.7}}$ & $\text{64.1}_{\pm \text{7.3}}$ & $\text{92.6}_{\pm \text{5.2}}$ & $\text{81.3}_{\pm \text{0.0}}$ & $\text{68.0}_{\pm \text{3.3}}$ & \hlsecond{$\text{84.7}_{\pm \text{3.9}}$} & $\text{81.9}_{\pm \text{3.2}}$ & \hlsecond{$\text{79.3}_{\pm \text{0.2}}$} & \hlsecond{$\text{66.7}_{\pm \text{0.5}}$} \\
                \quad StraTA (Ours) & \hlfirst{$\text{98.1}_{\pm \text{1.7}}$} & \hlsecond{$\text{71.8}_{\pm \text{8.9}}$} & \hlsecond{$\text{93.8}_{\pm \text{2.1}}$} & \hlfirst{$\text{97.9}_{\pm \text{3.6}}$} & \hlsecond{$\text{72.0}_{\pm \text{0.0}}$} & \hlfirst{$\text{88.9}_{\pm \text{2.4}}$} & \hlfirst{$\text{88.6}_{\pm \text{1.9}}$} & \hlfirst{$\text{85.9}_{\pm \text{0.5}}$} & \hlfirst{$\text{73.4}_{\pm \text{1.0}}$} \\
        \bottomrule
    \end{tabular}
\end{table}

\begin{figure}[ht]
    \centering
    \includegraphics[width=0.8\linewidth]{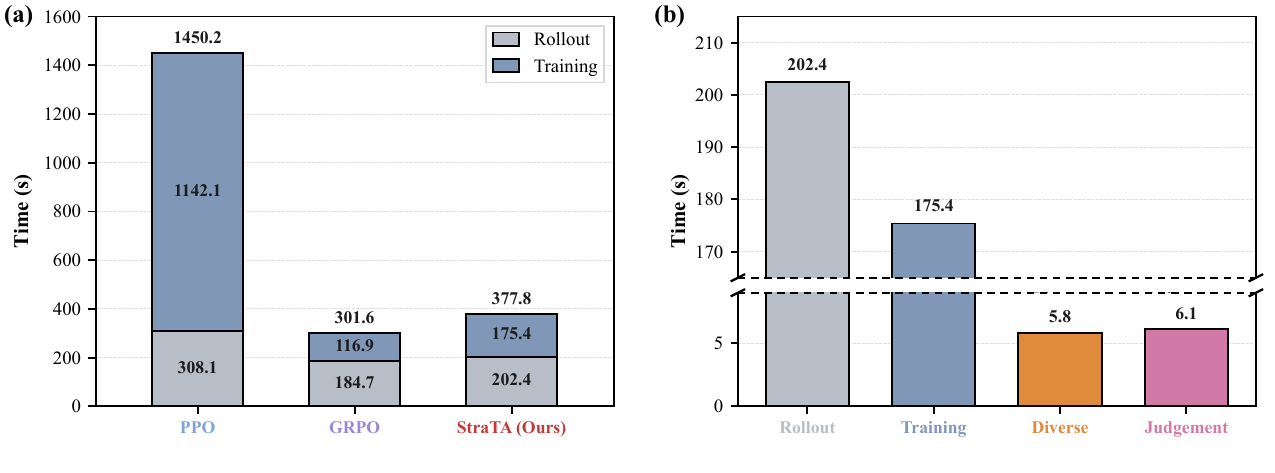}
    \vspace{-1.0em}
    \caption{Computational efficiency analysis on ALFWorld. (a) Per-step wall-clock time for PPO, GRPO, and StraTA, broken down into rollout and training phases. (b) Detailed per-step breakdown for StraTA, showing the time spent on rollout, training, and the two core techniques.}
    \label{fig:computational_efficiency}
\end{figure}

\noindent \textbf{Effectiveness of core components.} To assess the contribution of StraTA's key components, we conduct an ablation study using Qwen2.5-3B-Instruct~\citep{yang2024qwen} on ALFWorld and WebShop. To amplify the differences between variants, we reduce the batch size to 4 while keeping 150 training steps. We compare StraTA with three variants: (1) \textit{Vanilla}, the base version in Section~\ref{subsec:hierarchical_group_construction}; (2) \textit{Diverse}, which applies only diverse strategy rollout via farthest-point sampling in Section~\ref{subsec:diverse_strategy_rollout}; and (3) \textit{Judgment}, which applies only the step-level auxiliary reward via critical self-judgment in Section~\ref{subsec:step_level_auxiliary_reward}. Table~\ref{tab:ablation_components} reports the results. The \textit{Vanilla} variant already performs strongly. The \textit{Diverse} variant substantially improves ALFWorld performance to 87.9\%, while the \textit{Judgment} variant improves WebShop performance to 66.7\%. Figure~\ref{fig:ablation_components} intuitively demonstrates the convergence speed of different variants. These results indicate that the two components accelerate learning and improve performance in a largely orthogonal manner, jointly enabling StraTA's strong results on long-horizon agentic tasks.

\noindent \textbf{Analysis on computational efficiency.} We analyze the computational overhead of StraTA from both theoretical and empirical perspectives. For a given task, StraTA samples $N$ strategies, each paired with $M$ trajectories over a horizon of $H$, producing $N\times M\times H$ action rollouts and $N\times M$ judgment rollouts. Assuming that each rollout has similar latency and that $H$ is relatively large, the total computation is comparable to that of GRPO with a group size of $N\times M$. Moreover, the overhead introduced by the two core components accounts for only $1/H$ of the main action rollouts. Figure~\ref{fig:computational_efficiency}a shows the wall-clock time per training step for PPO, GRPO, and StraTA. StraTA is only marginally slower than GRPO in both rollout and training. Figure~\ref{fig:computational_efficiency}b further presents a stage-level cost breakdown for StraTA, showing that the overhead of the two components is negligible.

\section{Conclusion}
\label{sec:conclusion}

Long-horizon agentic RL remains difficult, which is largely due to the fact that purely reactive decision making weakens both exploration and credit assignment. To address this challenge, we introduce StraTA, a simple framework that equips agentic RL with explicit trajectory-level strategy. By generating a compact strategy from the initial task state and conditioning subsequent actions on it, StraTA separates high-level planning from low-level execution. Building on this formulation, hierarchical GRPO-style rollouts, diverse strategy rollout, and critical self-judgment jointly improve exploration and learning. Experiments on ALFWorld, WebShop, and SciWorld show that StraTA consistently improves both sample efficiency and final performance over strong prompting-based and RL baselines, underscoring the value of trajectory-level abstraction for long-horizon agents. StraTA nevertheless has limitations: its effectiveness depends on the quality of the generated strategy, and a fixed strategy may become restrictive when the environment changes substantially during an episode. Promising directions for future work include adaptive strategy revision, richer strategy representations, and extending the framework to broader agentic domains.

\section*{Acknowledgments and Disclosure of Funding}

This work was completed during Xiangyuan Xue's internship at Shanghai Artificial Intelligence Laboratory, and was supported by a locally commissioned task from the Shanghai Municipal Government. This work was also supported by the JC STEM Lab of AI for Science and Engineering, funded by The Hong Kong Jockey Club Charities Trust and the Research Grants Council of Hong Kong (Project No. CUHK14213224). No competing interests are declared for this work.

\bibliography{references}

\appendix

\section{Additional Experimental Results}
\label{appsec:additional_experimental_results}

\subsection{Effectiveness of Core Components}
\label{appsubsec:effectiveness_of_core_components}

\begin{figure}[ht]
    \centering
    \includegraphics[width=0.6\linewidth]{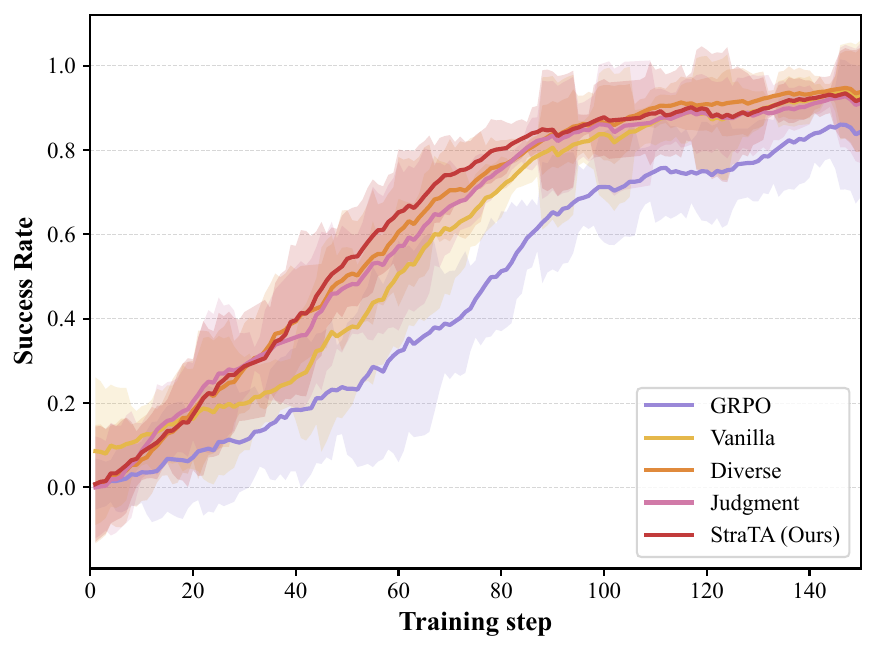}
    \caption{Training curves on ALFWorld for the ablation study on the effectiveness of core components. We compare StraTA with its ablated variants to evaluate the contribution of each component.}
    \label{fig:ablation_components}
\end{figure}

Section~\ref{subsec:ablation_studies} examines the effectiveness of StraTA's core components, and Table~\ref{tab:ablation_components} reports the numerical results for different variants. In this section, we present their training curves on ALFWorld~\citep{shridhar2021alfworld} in Figure~\ref{fig:ablation_components}. The \textit{Vanilla} variant already substantially outperforms the GRPO~\citep{shao2024deepseekmath} baseline, and adding the \textit{Diverse} and \textit{Judgment} techniques further accelerates convergence. Overall, StraTA with both techniques achieves the most stable and efficient training.

\subsection{Impact of Training Parameters}
\label{sec:ablation_hyperparams}

In this section, we further examine the impact of key hyperparameters, including the oversampling ratio $\sigma$, aggregation ratio $\delta$, and self-judgment reward weight $\kappa$. Following Section~\ref{subsec:ablation_studies}, we conduct ablation studies using Qwen2.5-3B-Instruct~\citep{yang2024qwen} on ALFWorld~\citep{shridhar2021alfworld} and WebShop~\citep{yao2022webshop}. The model is trained for 150 steps with a batch size of 4, while all other settings remain consistent with the main experiments.

Table~\ref{tab:ablation_sigma} reports the performance of StraTA with $\sigma$ set to $1$, $2$, $4$, and $8$. As $\sigma$ increases, performance improves steadily. This trend suggests that diverse strategy rollout effectively promotes exploration in the strategy space and yields more informative and discriminative reward signals.

Table~\ref{tab:ablation_delta} summarizes the results for $\delta$ values of $0.1$, $0.5$, and $1.0$. When $\delta$ is set to $0.1$, too few trajectories are used to reliably estimate strategy quality, resulting in poor performance. In contrast, setting $\delta$ to $1.0$ ignores potential execution errors during action selection, leading to overly pessimistic evaluations. Setting $\delta$ to $0.5$, as used in the main experiments, achieves the best performance.

Table~\ref{tab:ablation_kappa} shows the performance for $\kappa$ values of $0.01$, $0.1$, and $1.0$. A small weight of $0.01$ underutilizes the benefits of critical self-judgment, leading to degraded performance. Conversely, a large weight of $1.0$ places excessive reliance on self-judgment quality, which further harms performance. In general, setting $\kappa$ to $0.1$, adopted in the main experiments, provides the most effective trade-off.

\begin{table}[ht]
    \caption{Performance of StraTA on ALFWorld and WebShop across varying oversampling ratios $\sigma$. We sweep the values of $1$, $2$, $4$, and $8$ to evaluate its impact on performance.}
    \label{tab:ablation_sigma}
    \renewcommand{\arraystretch}{1.0}
    \centering \footnotesize \setlength{\tabcolsep}{3pt}
    \begin{tabular}{lccccccc|cc}
        \toprule
            \multicolumn{1}{c}{\multirow{2}{*}{\textbf{Parameter}}} & \multicolumn{7}{c|}{\textbf{ALFWorld}} & \multicolumn{2}{c}{\textbf{WebShop}} \\
            & Pick & Look & Clean & Heat & Cool & Pick2 & All & Score & Succ \\
        \midrule
            \multicolumn{10}{l}{\textit{Qwen2.5-3B-Instruct}} \\
                \quad $\sigma = 1$ & $\text{88.6}_{\pm \text{4.7}}$ & $\text{64.1}_{\pm \text{7.3}}$ & $\text{92.6}_{\pm \text{5.2}}$ & $\text{81.3}_{\pm \text{0.0}}$ & $\text{68.0}_{\pm \text{3.3}}$ & $\text{84.7}_{\pm \text{3.9}}$ & $\text{81.9}_{\pm \text{3.2}}$ & $\text{79.3}_{\pm \text{0.2}}$ & $\text{66.7}_{\pm \text{0.5}}$ \\
                \quad $\sigma = 2$ & $\text{93.3}_{\pm \text{1.4}}$ & \hlfirst{$\text{82.1}_{\pm \text{3.6}}$} & $\text{87.7}_{\pm \text{3.5}}$ & $\text{87.5}_{\pm \text{0.0}}$ & $\text{70.7}_{\pm \text{3.8}}$ & $\text{73.6}_{\pm \text{2.0}}$ & $\text{83.1}_{\pm \text{0.3}}$ & $\text{73.9}_{\pm \text{0.6}}$ & $\text{56.6}_{\pm \text{0.8}}$ \\
                \quad $\sigma = 4$ & $\text{92.4}_{\pm \text{1.3}}$ & $\text{61.5}_{\pm \text{0.0}}$ & \hlfirst{$\text{96.3}_{\pm \text{3.0}}$} & $\text{89.6}_{\pm \text{5.9}}$ & $\text{66.7}_{\pm \text{5.0}}$ & $\text{79.2}_{\pm \text{3.4}}$ & $\text{83.1}_{\pm \text{0.9}}$ & $\text{77.9}_{\pm \text{0.9}}$ & $\text{69.1}_{\pm \text{0.8}}$ \\
                \quad $\sigma = 8$ & \hlfirst{$\text{98.1}_{\pm \text{1.7}}$} & $\text{71.8}_{\pm \text{8.9}}$ & $\text{93.8}_{\pm \text{2.1}}$ & \hlfirst{$\text{97.9}_{\pm \text{3.6}}$} & \hlfirst{$\text{72.0}_{\pm \text{0.0}}$} & \hlfirst{$\text{88.9}_{\pm \text{2.4}}$} & \hlfirst{$\text{88.6}_{\pm \text{1.9}}$} & \hlfirst{$\text{85.9}_{\pm \text{0.5}}$} & \hlfirst{$\text{73.4}_{\pm \text{1.0}}$} \\
        \bottomrule
    \end{tabular}
\end{table}

\begin{table}[ht]
    \caption{Performance of StraTA on ALFWorld and WebShop across varying aggregation ratios $\delta$. We sweep the values of $0.1$, $0.5$, and $1.0$ to evaluate its impact on performance.}
    \label{tab:ablation_delta}
    \renewcommand{\arraystretch}{1.0}
    \centering \footnotesize \setlength{\tabcolsep}{3pt}
    \begin{tabular}{lccccccc|cc}
        \toprule
            \multicolumn{1}{c}{\multirow{2}{*}{\textbf{Parameter}}} & \multicolumn{7}{c|}{\textbf{ALFWorld}} & \multicolumn{2}{c}{\textbf{WebShop}} \\
            & Pick & Look & Clean & Heat & Cool & Pick2 & All & Score & Succ \\
        \midrule
            \multicolumn{10}{l}{\textit{Qwen2.5-3B-Instruct}} \\
                \quad $\delta = 0.1$ & $\text{89.5}_{\pm \text{2.7}}$ & \hlfirst{$\text{71.8}_{\pm \text{3.6}}$} & $\text{82.7}_{\pm \text{3.5}}$ & $\text{50.0}_{\pm \text{17.7}}$ & $\text{34.7}_{\pm \text{5.0}}$ & $\text{81.9}_{\pm \text{2.0}}$ & $\text{71.0}_{\pm \text{0.7}}$ & $\text{79.3}_{\pm \text{0.3}}$ & $\text{66.1}_{\pm \text{0.3}}$ \\
                \quad $\delta = 0.5$ & \hlfirst{$\text{98.1}_{\pm \text{1.7}}$} & \hlfirst{$\text{71.8}_{\pm \text{8.9}}$} & \hlfirst{$\text{93.8}_{\pm \text{2.1}}$} & \hlfirst{$\text{97.9}_{\pm \text{3.6}}$} & $\text{72.0}_{\pm \text{0.0}}$ & \hlfirst{$\text{88.9}_{\pm \text{2.4}}$} & \hlfirst{$\text{88.6}_{\pm \text{1.9}}$} & \hlfirst{$\text{85.9}_{\pm \text{0.5}}$} & \hlfirst{$\text{73.4}_{\pm \text{1.0}}$} \\
                \quad $\delta = 1.0$ & $\text{92.4}_{\pm \text{1.4}}$ & $\text{69.2}_{\pm \text{6.3}}$ & $\text{92.6}_{\pm \text{3.0}}$ & $\text{85.4}_{\pm \text{5.9}}$ & \hlfirst{$\text{82.7}_{\pm \text{1.9}}$} & $\text{76.4}_{\pm \text{2.0}}$ & $\text{85.0}_{\pm \text{0.6}}$ & $\text{85.2}_{\pm \text{0.7}}$ & $\text{71.3}_{\pm \text{1.5}}$ \\
        \bottomrule
    \end{tabular}
\end{table}

\begin{table}[ht]
    \caption{Performance of StraTA on ALFWorld and WebShop across varying self-judgment reward weights $\kappa$. We sweep the values of $0.01$, $0.1$, and $1.0$ to evaluate its impact on performance.}
    \label{tab:ablation_kappa}
    \renewcommand{\arraystretch}{1.0}
    \centering \footnotesize \setlength{\tabcolsep}{3pt}
    \begin{tabular}{lccccccc|cc}
        \toprule
            \multicolumn{1}{c}{\multirow{2}{*}{\textbf{Parameter}}} & \multicolumn{7}{c|}{\textbf{ALFWorld}} & \multicolumn{2}{c}{\textbf{WebShop}} \\
            & Pick & Look & Clean & Heat & Cool & Pick2 & All & Score & Succ \\
        \midrule
            \multicolumn{10}{l}{\textit{Qwen2.5-3B-Instruct}} \\
                \quad $\kappa = 0.01$ & $\text{94.3}_{\pm \text{0.0}}$ & \hlfirst{$\text{71.8}_{\pm \text{3.6}}$} & $\text{88.9}_{\pm \text{0.0}}$ & $\text{87.5}_{\pm \text{0.0}}$ & $\text{72.0}_{\pm \text{0.0}}$ & $\text{83.3}_{\pm \text{0.0}}$ & $\text{84.5}_{\pm \text{0.3}}$ & $\text{81.0}_{\pm \text{0.4}}$ & $\text{73.3}_{\pm \text{0.8}}$ \\
                \quad $\kappa = 0.1$ & \hlfirst{$\text{98.1}_{\pm \text{1.7}}$} & \hlfirst{$\text{71.8}_{\pm \text{8.9}}$} & $\text{93.8}_{\pm \text{2.1}}$ & \hlfirst{$\text{97.9}_{\pm \text{3.6}}$} & $\text{72.0}_{\pm \text{0.0}}$ & \hlfirst{$\text{88.9}_{\pm \text{2.4}}$} & \hlfirst{$\text{88.6}_{\pm \text{1.9}}$} & \hlfirst{$\text{85.9}_{\pm \text{0.5}}$} & \hlfirst{$\text{73.4}_{\pm \text{1.0}}$} \\
                \quad $\kappa = 1.0$ & $\text{94.3}_{\pm \text{0.0}}$ & $\text{61.5}_{\pm \text{0.0}}$ & \hlfirst{$\text{96.3}_{\pm \text{0.0}}$} & $\text{93.8}_{\pm \text{5.1}}$ & \hlfirst{$\text{76.0}_{\pm \text{0.0}}$} & $\text{87.5}_{\pm \text{0.0}}$ & $\text{87.1}_{\pm \text{0.6}}$ & $\text{78.9}_{\pm \text{0.8}}$ & $\text{61.8}_{\pm \text{1.0}}$ \\
        \bottomrule
    \end{tabular}
\end{table}

\subsection{Reliability of Critical Self-judgment}
\label{appsubsec:reliability_of_critical_self_judgment}

Section~\ref{subsec:step_level_auxiliary_reward} introduces critical self-judgment as a step-level auxiliary reward, where the reward signal is constructed by the agent itself rather than by an external supervisor. Such a self-generated signal is not guaranteed to be reliable. We therefore verify its reliability with GPT-5.5 as an independent judge, following LLM-as-a-judge techniques~\citep{zheng2023judging,gu2024survey,li2025generation}. The verification covers two perspectives: the accuracy of the self-judgment results, and whether the mechanism induces reward hacking through superficial compliance.

We first examine how faithfully the agent identifies its own problematic steps. Table~\ref{tab:judgment_reliability} compares StraTA's step-level judgments with GPT-5.5 under the same judgment protocol. From the 0th to the 50th training step, accuracy increases from 72.6\% to 90.1\%, while F1 increases from 77.1\% to 84.9\%, indicating that the reliability of self-judgment improves during training.

We then examine whether the self-generated reward encourages the agent to appear to follow the strategy without making genuine task progress. We use GPT-5.5 to independently audit 200 trajectories from each of ALFWorld and WebShop without access to the self-judgment labels. Each trajectory is classified as fully reliable, suspected deception, or confirmed deception. Table~\ref{tab:reward_hacking} reports that most of the trajectories are fully reliable, with no confirmed deception observed.

Overall, these results indicate that critical self-judgment is reliable in practice. A certain fraction of misjudgments does remain, but it steadily decreases as training proceeds. Moreover, the resulting reward carries a small weight $\kappa$, so it refines rather than overrides the reliable supervision provided by the outcome reward, and thus accelerates convergence.

\begin{table}[ht]
    \caption{Reliability of critical self-judgment. We compare StraTA's step-level judgments with GPT-5.5 under the same judgment protocol using 200 trajectories at the 0th and 50th training steps.}
    \label{tab:judgment_reliability}
    \renewcommand{\arraystretch}{1.1}
    \centering \footnotesize \setlength{\tabcolsep}{5pt}
    \begin{tabular}{lcccccc}
        \toprule
            \textbf{Checkpoint} & \textbf{\#Trajectories} & \textbf{\#Steps} & \textbf{Accuracy} & \textbf{Precision} & \textbf{Recall} & \textbf{F1-score} \\
        \midrule
            Step 0 & 200 & 2,640 & 72.6\% & 80.3\% & 74.1\% & 77.1\% \\
            Step 50 & 200 & 1,944 & 90.1\% & 80.9\% & 89.2\% & 84.9\% \\
        \bottomrule
    \end{tabular}
\end{table}

\begin{table}[ht]
    \caption{Audit of reward hacking. GPT-5.5 inspects 200 trajectories per environment, and classifies each trajectory as fully reliable, suspected deception, or confirmed deception.}
    \label{tab:reward_hacking}
    \renewcommand{\arraystretch}{1.1}
    \centering \footnotesize \setlength{\tabcolsep}{5pt}
    \begin{tabular}{lcccc}
        \toprule
            \textbf{Environment} & \textbf{\#Trajectories} & \textbf{Fully Reliable} & \textbf{Suspected Deception} & \textbf{Confirmed Deception} \\
        \midrule
            ALFWorld & 200 & 98.5\% & 1.5\% & 0.0\% \\
            WebShop & 200 & 97.0\% & 3.0\% & 0.0\% \\
        \bottomrule
    \end{tabular}
\end{table}

\section{Additional Implementation Details}
\label{appsec:additional_implementation_details}

\subsection{Environmental Setup}
\label{appsubsec:environmental_setup}

All benchmark environments are accessed through AgentGym~\citep{xi2025agentgym}, which provides a unified interface for both training and evaluation. This setup standardizes the observation format, action interface, interaction loop, and reward computation across ALFWorld~\citep{shridhar2021alfworld}, WebShop~\citep{yao2022webshop}, and SciWorld~\citep{wang2022scienceworld}.

\noindent \textbf{ALFWorld.} ALFWorld~\citep{shridhar2021alfworld} is a text-based embodied household environment in which an agent interacts with simulated home scenes through natural-language actions. Each task requires the agent to complete a goal-oriented household instruction, such as picking up an object, examining an object under a specified condition, cleaning an object, heating or cooling an object, or placing two objects in the same location. We report the overall success rate as well as the success rates for six task types: Pick, Look, Clean, Heat, Cool, and Pick2. Each episode allows up to 50 environment steps. During training, the environment provides a rule-based, step-decayed reward, with a maximum value of $1$ for successful completion and $0$ for failure. This design preserves the binary success signal while assigning larger rewards to trajectories that complete the task in fewer steps. We also apply a format penalty of $-1$ to malformed responses that cannot be parsed into the required action format.

\noindent \textbf{WebShop.} WebShop~\citep{yao2022webshop} is a web-based shopping environment that evaluates an agent's ability to complete realistic purchasing tasks through multi-step web interaction. Given a user instruction, the agent must search for relevant products, inspect product pages, compare attributes such as price and specifications, and select an item that satisfies the task requirements. Each episode is limited to 15 environment steps. Following the standard evaluation protocol, we report both the task score and the success rate. The task score measures partial progress toward the target product, whereas the success rate indicates whether the final selected product fully satisfies the user instruction. During training, the environment provides a rule-based, step-decayed reward, with a maximum value of $1$ for success and $0$ for failure. This design preserves the binary success signal while assigning larger rewards to trajectories that complete the task in fewer steps. A format penalty of $-1$ is applied to malformed responses that cannot be parsed into the required action format.

\noindent \textbf{SciWorld.} SciWorld~\citep{wang2022scienceworld} is a text-based scientific experimentation environment that requires the agent to perform procedural reasoning and execute actions in simulated science tasks. The benchmark covers multiple task categories, including but not limited to measurement, conditional testing, object finding, chemical mixing, and lifespan reasoning. Following the standard benchmark setting, we report the normalized task score as the primary evaluation metric. Each episode allows up to 20 environment steps. During training, the reward is computed from the final task score, normalized to the range $[0, 1]$, and further adjusted by a step-decay factor so that more efficient task completion receives a higher reward. As in the other environments, malformed responses that cannot be parsed into the required action format receive a format penalty of $-1$.

\subsection{Prompt Templates}
\label{appsubsec:prompt_templates}

Section~\ref{sec:method} introduces the StraTA pipeline, and Figure~\ref{fig:method} illustrates the sampling process. For reproducibility, we provide the prompt templates used for global strategy generation, strategy-guided action selection, and critical self-judgment in this section.

\begin{tcolorbox}[title=A Template Prompt for Global Strategy Generation, breakable, width=\textwidth]
\begin{Verbatim}[breaklines=true, breakanywhere=true, formatcom=\normalfont, fontsize=\small]
{description}

{context}

Before you start to work on the task, you should propose a global strategy. The requirements are presented as follows:
1. The strategy is a constant text snippet, which guides your future actions throughout the whole episode to achieve the goal.
2. The strategy should be concrete enough, so that future actions can be taken by strictly following it.
3. The strategy should be practical enough, so that future actions can be taken based on existing information.

First think step by step to come up with an overall planning. Then propose your strategy enclosed within <strategy>...</strategy> tags. For example:
According to the rules, any non-random strategy will lead to sub-optimal outcomes.
<strategy>Select one of the actions uniformly at random at each step.</strategy>
\end{Verbatim}
\end{tcolorbox}

\begin{tcolorbox}[title=A Template Prompt for Strategy-guided Action Taking, breakable, width=\textwidth]
\begin{Verbatim}[breaklines=true, breakanywhere=true, formatcom=\normalfont, fontsize=\small]
{description}

You should strictly observe the following strategy to take actions:
{strategy}

{context}

{instruction}
\end{Verbatim}
\end{tcolorbox}

\begin{tcolorbox}[title=A Template Prompt for Critical Self-judgment, breakable, width=\textwidth]
\begin{Verbatim}[breaklines=true, breakanywhere=true, formatcom=\normalfont, fontsize=\small]
{description}

The proposed global strategy for the task is presented as follows:
{strategy}

The complete history of actions taken is presented as follows:
{history}

You should point out all the problematic steps. A step is said to be problematic if it neither follows the global strategy nor pushes forward the task progress.

First think step by step to analyze the history. Then provide your judgment enclosed within <judgment>...</judgment> tags. The content should be a list of step numbers corresponding to the given history. If all the steps are reasonable, yield an empty list. For example:
Step 1 seems to deviate from the strategy, but it effectively facilitates future exploration. Step 2 and 5 fail to follow the strategy and make no contribution to the final success.
<judgment>[2, 5]</judgment>
\end{Verbatim}
\end{tcolorbox}

\subsection{Training Parameters}
\label{appsubsec:training_parameters}

In this section, we provide additional training details that complement the setup described in Section~\ref{subsec:experimental_setup}. Our implementation of StraTA is mainly built on the rLLM framework~\citep{rllm2025}. The complete parameter settings are summarized in Table~\ref{tab:training_parameters}. Most of these parameters follow GiGPO~\citep{feng2025group}, including the AdamW optimizer~\citep{loshchilov2017decoupled} and the asymmetric clipping bounds $\epsilon_{\text{low}}$ and $\epsilon_{\text{high}}$ of the PPO objective introduced by DAPO~\citep{yu2025dapo}.

\begin{table}[ht]
    \centering
    \caption{The detailed parameter settings when implementing our StraTA framework.}
    \label{tab:training_parameters}
    \begin{tabular}{lc}
        \toprule
            \textbf{Parameter} & \textbf{Setting} \\
        \midrule
            Base model & Qwen2.5-7B-Instruct \\
            Number of strategies per task $N$ & 4 \\
            Number of rollouts per strategy $M$ & 8 \\
            Oversampling ratio $\sigma$ & 8 \\
            Aggregation ratio $\delta$ & 0.5 \\
            Length penalty threshold $\lambda$ & 0.5 \\
            Self-judgment reward weight $\kappa$ & 0.1 \\
            Embedding model & MiniLM-L6 \\
            Token limit for prompts & 7168 \\
            Token limit for responses & 1024 \\
            Training temperature & 1.0 \\
            Evaluation temperature & 0.7 \\
            Evaluation top-p & 0.8 \\
            Evaluation top-k & 20 \\
            Clipping $\epsilon_{\text{low}}$ & 0.2 \\
            Clipping $\epsilon_{\text{high}}$ & 0.28 \\
            Optimizer & AdamW \\
            Learning rate  & 1e-6 \\
            Weight decay & 0.01 \\
            Mini batch size & 1024 \\
            Number of training steps & 150 \\
        \bottomrule
    \end{tabular}
\end{table}

\subsection{Diverse Strategy Rollout via Farthest Point Sampling}
\label{appsubsec:diverse_strategy_rollout}

Section~\ref{subsec:diverse_strategy_rollout} introduces diverse strategy rollout, in which farthest point sampling in semantic space is used to select diverse strategies. Here, we provide additional details of this technique. Algorithm~\ref{alg:diverse_strategy_rollout} summarizes the procedure. The overall complexity is $O(\sigma N^2 d)$, where $d$ denotes the dimensionality of the semantic embedding. In practice, selection takes only a few milliseconds per task, which is negligible relative to the cost of LLM rollouts, as further confirmed in Section~\ref{subsec:ablation_studies}.

\begin{algorithm}[ht]
    \caption{Diverse Strategy Selection via Farthest Point Sampling}
    \label{alg:diverse_strategy_rollout}
    \begin{algorithmic}[1]
        \Require Oversampled candidates $\mathcal{C}=\{z^i\}_{i=1}^{\sigma N}$ and semantic encoder $\phi(\cdot)$
        \Ensure Diverse strategies $\mathcal{Z}=\{z^i\}_{i=1}^{N}$
        \For{each candidate strategy $z^i \in \mathcal{C}$}
            \State $e^i \leftarrow \phi(z^i)$
            \State $e^i \leftarrow e^i / \|e^i\|_2$
        \EndFor
        \State $\bar{e} \leftarrow \frac{1}{|\mathcal{C}|} \sum_{z^i \in \mathcal{C}} e^i$
        \State $z^* \leftarrow \arg\max_{z^i \in \mathcal{C}} {e^i}^{\top}\bar{e}$
        \State $\mathcal{Z} \leftarrow \{z^*\}$
        \While{$|\mathcal{Z}| < N$}
            \State $z_{\mathrm{next}} \leftarrow \arg\min_{z^i \in \mathcal{C}\setminus\mathcal{Z}} \max_{z^j \in \mathcal{Z}} {e^i}^{\top} e^j$
            \State $\mathcal{Z} \leftarrow \mathcal{Z} \cup \{z_{\mathrm{next}}\}$
        \EndWhile
        \State \Return $\mathcal{Z}$
    \end{algorithmic}
\end{algorithm}

\section{Case Studies}
\label{appsec:case_studies}

To further illustrate the behavior of StraTA, we provide case studies in this section, including one example each for ALFWorld~\citep{shridhar2021alfworld}, WebShop~\citep{yao2022webshop}, and SciWorld~\citep{wang2022scienceworld}. All the examples are randomly selected from the evaluation history of the models trained in the main experiments.

\subsection{ALFWorld}

\definecolor{deeppurple}{RGB}{96,40,140}

\newcommand{\alfworldenvbox}[2]{%
    \noindent
    \begin{minipage}{0.8\linewidth}
    \centering
    \resizebox{0.9\linewidth}{!}{%
    \begin{tcolorbox}[
        colback=gray!5!white,
        colframe=orange!75!black,
        title=#1,
        boxrule=0.3mm,
        width=\textwidth,
        arc=1mm,
        auto outer arc=true
    ]
    #2
    \end{tcolorbox}%
    }
    \end{minipage}\par
}

\newcommand{\alfworldagentbox}[2]{%
    \hfill
    \begin{minipage}{0.8\linewidth}
    \centering
    \resizebox{0.9\linewidth}{!}{%
    \begin{tcolorbox}[
        colback=gray!5!white,
        colframe=blue!75!black,
        title=#1,
        boxrule=0.3mm,
        width=\textwidth,
        arc=1mm,
        auto outer arc=true
    ]
    #2
    \end{tcolorbox}%
    }
    \end{minipage}\par
}

\newcommand{\alfworldexampletag}[3]{%
    \textcolor{#1}{<#2>}#3\textcolor{#1}{</#2>}%
}

\alfworldenvbox{Environment (Initial Observation)}{%
Task: put a tissuebox in sidetable.

You are in the middle of a room. Looking quickly around you, you see an armchair 1, a coffeetable 1, a drawer 4, a drawer 3, a drawer 2, a drawer 1, a dresser 1, a garbagecan 1, a sidetable 3, a sidetable 2, a sidetable 1, and a sofa 1.
}

\alfworldagentbox{Agent (Strategy Generation)}{%
\alfworldexampletag{deeppurple}{strategy}{Explore the room to find where the tissuebox might be and how to get it to the sidetable.}
}

\alfworldenvbox{Environment (Step 1)}{%
You are in the middle of a room. Looking quickly around you, you see an armchair 1, a coffeetable 1, a drawer 4, a drawer 3, a drawer 2, a drawer 1, a dresser 1, a garbagecan 1, a sidetable 3, a sidetable 2, a sidetable 1, and a sofa 1.
}

\alfworldagentbox{Agent (Step 1)}{%
\alfworldexampletag{deeppurple}{action}{go to coffeetable 1}
}

\alfworldenvbox{Environment (Step 2)}{%
On the coffeetable 1, you see a box 1, a cellphone 3, a pencil 1, a statue 1, a tissuebox 2, and a tissuebox 1.
}

\alfworldagentbox{Agent (Step 2)}{%
\alfworldexampletag{deeppurple}{action}{take tissuebox 1 from coffeetable 1}
}

\alfworldenvbox{Environment (Step 3)}{%
You pick up the tissuebox 1 from the coffeetable 1.
}

\alfworldagentbox{Agent (Step 3)}{%
\alfworldexampletag{deeppurple}{action}{go to sidetable 1}
}

\alfworldenvbox{Environment (Step 4)}{%
On the sidetable 1, you see a cellphone 2 and a houseplant 1.
}

\alfworldagentbox{Agent (Step 4)}{%
\alfworldexampletag{deeppurple}{action}{put tissuebox 1 in/on sidetable 1}
}

\alfworldenvbox{Environment (Final Feedback)}{%
You put the tissuebox 1 in/on the sidetable 1. The task is completed successfully.
}

\subsection{WebShop}

\definecolor{deeppurple}{RGB}{96,40,140}

\newcommand{\webshopenvbox}[2]{%
    \noindent
    \begin{minipage}{0.8\linewidth}
    \centering
    \resizebox{0.9\linewidth}{!}{%
    \begin{tcolorbox}[
        colback=gray!5!white,
        colframe=orange!75!black,
        title=#1,
        boxrule=0.3mm,
        width=\textwidth,
        arc=1mm,
        auto outer arc=true
    ]
    #2
    \end{tcolorbox}%
    }
    \end{minipage}\par
}

\newcommand{\webshopagentbox}[2]{%
    \hfill
    \begin{minipage}{0.8\linewidth}
    \centering
    \resizebox{0.9\linewidth}{!}{%
    \begin{tcolorbox}[
        colback=gray!5!white,
        colframe=blue!75!black,
        title=#1,
        boxrule=0.3mm,
        width=\textwidth,
        arc=1mm,
        auto outer arc=true
    ]
    #2
    \end{tcolorbox}%
    }
    \end{minipage}\par
}

\newcommand{\webshopexampletag}[3]{%
    \textcolor{#1}{<#2>}#3\textcolor{#1}{</#2>}%
}

\webshopenvbox{Environment (Initial Observation)}{%
Task: Find me high density, ready use, twin size, long lasting, assembly required mattresses with box spring, and price lower than 550.00 dollars.
}

\webshopagentbox{Agent (Strategy Generation)}{%
\webshopexampletag{deeppurple}{strategy}{Start with a generic search query to identify available products fitting the criteria. Then, refine the search based on user feedback to find the best option that meets the specifications.}
}

\webshopenvbox{Environment (Step 1)}{%
Search
}

\webshopagentbox{Agent (Step 1)}{%
\webshopexampletag{deeppurple}{action}{search[High density, ready use, twin size, long lasting, assembly required mattresses with box spring, price lower than 550]}
}

\webshopenvbox{Environment (Step 2)}{%
Back to Search | Page 1 (Total results: 50) | Next >

B078PKLZFB | Mattress Solution, 13-Inch Soft Foam Encased Hybrid Eurotop Pillowtop Innerspring Mattress And Split Wood Traditional Box Spring/Foundation Set With Frame, Twin Size 74" x 38" | \$520.0

B07SXLVPF3 | Mayton 12-Inch King Size Mattress, Box Spring And Bed Frame - Foam Encased Soft Pillow Top Hybrid Contouring Comfort, No Assembly Required 78x79 | \$884.03

B01A0L5FXA | Spring Coil 1 Mattress, Queen | \$565.71

B09H3N5P74 | Foamma 3" x 24" x 72" Mattress for RV with Water Resistant Organic Cotton Cover, Firm High Density Foam, USA Made, CertiPUR-US Certified Foam | \$149.99
}

\webshopagentbox{Agent (Step 2)}{%
\webshopexampletag{deeppurple}{action}{click[b078pklzfb]}
}

\webshopenvbox{Environment (Step 3)}{%
Back to Search | < Prev

Mattress Solution, 13-Inch Soft Foam Encased Hybrid Eurotop Pillowtop Innerspring Mattress And Split Wood Traditional Box Spring/Foundation Set With Frame, Twin Size 74" x 38"

Price: \$520.0 | Rating: N.A. | Description | Features | Reviews | Buy Now
}

\webshopagentbox{Agent (Step 3)}{%
\webshopexampletag{deeppurple}{action}{click[buy now]}
}

\webshopenvbox{Environment (Final Feedback)}{%
Thank you for shopping with us! | Your code: | ... | Your score (min 0.0, max 1.0) | 1.0 | Reward Details | None

The task is completed successfully with a score of $1.0$.
}

\subsection{SciWorld}

\definecolor{deeppurple}{RGB}{96,40,140}
\newcommand{\sciworldenvbox}[2]{%
    \noindent
    \begin{minipage}{0.8\linewidth}
    \centering
    \resizebox{0.9\linewidth}{!}{%
    \begin{tcolorbox}[
        colback=gray!5!white,
        colframe=orange!75!black,
        title=#1,
        boxrule=0.3mm,
        width=\textwidth,
        arc=1mm,
        auto outer arc=true
    ]
    #2
    \end{tcolorbox}%
    }
    \end{minipage}\par
}
\newcommand{\sciworldagentbox}[2]{%
    \hfill
    \begin{minipage}{0.8\linewidth}
    \centering
    \resizebox{0.9\linewidth}{!}{%
    \begin{tcolorbox}[
        colback=gray!5!white,
        colframe=blue!75!black,
        title=#1,
        boxrule=0.3mm,
        width=\textwidth,
        arc=1mm,
        auto outer arc=true
    ]
    #2
    \end{tcolorbox}%
    }
    \end{minipage}\par
}
\newcommand{\sciworldexampletag}[3]{%
    \textcolor{#1}{<#2>}#3\textcolor{#1}{</#2>}%
}

\sciworldenvbox{Environment (Initial Observation)}{%
Task: Your task is to find the animal with the longest life span. The animals are in the `outside' location. Focus on the animal with the longest life span.\\
This room is called the living room. In it, you see:\\
\hspace*{1em} the agent\\
\hspace*{1em} a substance called air\\
\hspace*{1em} a chair. On the chair is: nothing.\\
\hspace*{1em} a couch. On the couch is: a white pillow.\\
\hspace*{1em} a desk. On the desk is: a drawer.\\
\hspace*{1em} a painting\\
You also see:\\
\hspace*{1em} A door to the hallway (that is closed)
}

\sciworldagentbox{Agent (Strategy Generation)}{%
\sciworldexampletag{deeppurple}{strategy}{To achieve this goal, the first step is to open the hallway door to gain visibility of the hallway. This can be done by moving towards and then opening the hallway door. After observing this new environment, you can then identify and compare the life spans of any animals you encounter outside.}
}

\sciworldenvbox{Environment (Step 1)}{%
This room is called the living room. In it, you see:\\
\hspace*{1em} the agent\\
\hspace*{1em} a substance called air\\
\hspace*{1em} a chair. On the chair is: nothing.\\
\hspace*{1em} a couch. On the couch is: a white pillow.\\
\hspace*{1em} a desk. On the desk is: a drawer.\\
\hspace*{1em} a painting\\
You also see:\\
\hspace*{1em} A door to the hallway (that is closed)
}

\sciworldagentbox{Agent (Step 1)}{%
\sciworldexampletag{deeppurple}{action}{open("door to hallway")}
}

\sciworldenvbox{Environment (Step 2)}{%
The door is now open.
}

\sciworldagentbox{Agent (Step 2)}{%
\sciworldexampletag{deeppurple}{action}{goto("hallway")}
}

\sciworldenvbox{Environment (Step 3)}{%
You move to the hallway.
}

\sciworldagentbox{Agent (Step 3)}{%
\sciworldexampletag{deeppurple}{action}{lookaround()}
}

\sciworldenvbox{Environment (Step 4)}{%
This room is called the hallway. In it, you see:\\
\hspace*{1em} the agent\\
\hspace*{1em} a substance called air\\
\hspace*{1em} a finger painting\\
You also see:\\
\hspace*{1em} A door to the art studio (that is closed)\\
\hspace*{1em} A door to the bedroom (that is closed)\\
\hspace*{1em} A door to the greenhouse (that is closed)\\
\hspace*{1em} A door to the kitchen (that is closed)\\
\hspace*{1em} A door to the living room (that is open)\\
\hspace*{1em} A door to the workshop (that is closed)
}

\sciworldagentbox{Agent (Step 4)}{%
\sciworldexampletag{deeppurple}{action}{open("door to greenhouse")}
}

\sciworldenvbox{Environment (Step 5)}{%
The door is now open.
}

\sciworldagentbox{Agent (Step 5)}{%
\sciworldexampletag{deeppurple}{action}{goto("greenhouse")}
}

\sciworldenvbox{Environment (Step 6)}{%
You move to the greenhouse.
}

\sciworldagentbox{Agent (Step 6)}{%
\sciworldexampletag{deeppurple}{action}{lookaround()}
}

\sciworldenvbox{Environment (Step 7)}{%
This room is called the greenhouse. In it, you see:\\
\hspace*{1em} the agent\\
\hspace*{1em} a substance called air\\
\hspace*{1em} a bee hive. The bee hive door is closed.\\
\hspace*{1em} a jug (containing nothing)\\
\hspace*{1em} a sink, which is turned off. In the sink is: nothing.\\
You also see:\\
\hspace*{1em} A door to the hallway (that is open)\\
\hspace*{1em} A door to the outside (that is closed)
}

\sciworldagentbox{Agent (Step 7)}{%
\sciworldexampletag{deeppurple}{action}{open("door to outside")}
}

\sciworldenvbox{Environment (Step 8)}{%
The door is now open.
}

\sciworldagentbox{Agent (Step 8)}{%
\sciworldexampletag{deeppurple}{action}{goto("outside")}
}

\sciworldenvbox{Environment (Step 9)}{%
You move to the outside.
}

\sciworldagentbox{Agent (Step 9)}{%
\sciworldexampletag{deeppurple}{action}{lookaround()}
}

\sciworldenvbox{Environment (Step 10)}{%
This outside location is called the outside. Here you see:\\
\hspace*{1em} the agent\\
\hspace*{1em} a substance called air\\
\hspace*{1em} an axe\\
\hspace*{1em} a frog egg\\
\hspace*{1em} a baby elephant\\
\hspace*{1em} a fire pit (containing nothing)\\
\hspace*{1em} a fountain (containing a substance called water)\\
\hspace*{1em} the ground\\
\hspace*{1em} a baby hedgehog\\
\hspace*{1em} a shovel\\
\hspace*{1em} a substance called wood\\
You also see:\\
\hspace*{1em} A door to the foundry (that is closed)\\
\hspace*{1em} A door to the greenhouse (that is open)\\
\hspace*{1em} A door to the kitchen (that is closed)
}

\sciworldagentbox{Agent (Step 10)}{%
\sciworldexampletag{deeppurple}{action}{focus("baby elephant")}
}

\sciworldenvbox{Environment (Final Feedback)}{%
The task is completed successfully.
}

\end{document}